%% file: mainbody.tex
\documentclass[10pt,journal,compsoc]{IEEEtran}

\usepackage{dutchcal}
\usepackage{graphicx}
\usepackage[countmax]{subfloat}
\usepackage[caption=false,font=normalsize,labelfont=sf,textfont=sf]{subfig}
\usepackage{amsmath,amsfonts,amssymb}
\usepackage{color}
\usepackage{algorithmic}
\usepackage{algorithm}
\usepackage{multirow,booktabs}
\usepackage{xcolor}
\usepackage[mathscr]{euscript}
\usepackage{pifont}
\usepackage{bbding}

\newtheorem{Proposition}{Proposition}

\ifCLASSOPTIONcompsoc
  \usepackage{cite}
\else
  \usepackage{cite}
\fi
\ifCLASSINFOpdf
\else
\fi
\newcommand\MYhyperrefoptions{bookmarks=true,bookmarksnumbered=true,
pdfpagemode={UseOutlines},plainpages=false,pdfpagelabels=true,
colorlinks=true,linkcolor={black},citecolor={black},urlcolor={black},
pdftitle={Bare Demo of IEEEtran.cls for Computer Society Journals},
pdfsubject={Typesetting},
pdfauthor={Michael D. Shell},
pdfkeywords={Computer Society, IEEEtran, journal, LaTeX, paper,
             template}}
\ifCLASSINFOpdf
\usepackage[\MYhyperrefoptions,pdftex]{hyperref}
\else
\usepackage[\MYhyperrefoptions,breaklinks=true,dvips]{hyperref}
\usepackage{breakurl}
\fi
\begin{document}
%
    \title{Selection, Ensemble, and Adaptation: Advancing Multi-Source-Free Domain Adaptation via Architecture Zoo}
%
%
%
%

\author{
Jiangbo~Pei,
Ruizhe~Li,
Aidong~Men,
Yang~Liu,
Xiahai Zhuang,
and~Qingchao~Chen\textsuperscript{\ding{41}}
\IEEEcompsocitemizethanks{\IEEEcompsocthanksitem Jiangbo Pei is affiliated with the National Institute of Health Data Science, Peking University, Beijing, 100191, China.
He is also with the School of Artificial Intelligence, Beijing University of Posts and Telecommunications, Beijing 100876, China.  (e-mail: jiangbop@bupt.edu.cn).
\IEEEcompsocthanksitem
Ruizhe Li and Aidong Men are with the School of Artificial Intelligence, Beijing University of Posts and Telecommunications, Beijing 100876, China.  (e-mail: liruizhe@bupt.edu.cn; menad@bupt.edu.cn).
\IEEEcompsocthanksitem Yang Liu is with Wangxuan Institute of Computer Technology at Peking University,  Beijing, 100080, China. (email: yangliu@pku.edu.cn).
\IEEEcompsocthanksitem Xiahai Zhuang is with the School of Data Science, Fudan University, Shanghai, 200433, China. (e-mail: zxh@fudan.edu.cn).
\IEEEcompsocthanksitem Qingchao Chen is with the National Institute of Health Data Science, Peking University, Beijing, 100191, China. (e-mail: qingchao.chen@pku.edu.cn).}
\thanks{\ding{41} Corresponding author.}
}

%
%

\markboth{Journal of \LaTeX\ Class Files,~Vol.~14, No.~8, August~2015}%
{Shell \MakeLowercase{\textit{et al.}}: Bare Advanced Demo of IEEEtran.cls for IEEE Computer Society Journals}
%



\IEEEtitleabstractindextext{%
\begin{abstract}
Conventional Multi-Source Free Domain Adaptation (MSFDA) assumes that each source domain provides a single source model, and all source models adopt a uniform architecture. This paper introduces Zoo-MSFDA, a more general setting that allows each source domain to offer a zoo of multiple source models with different architectures. While it enriches the source knowledge, Zoo-MSFDA risks being dominated by suboptimal/harmful models. To address this issue, we theoretically analyze the model selection problem in Zoo-MSFDA, and introduce two principles: \textit{transferability principle} and \textit{diversity principle}. Recognizing the challenge of measuring transferability, we subsequently propose a novel Source-Free Unsupervised Transferability Estimation (SUTE). It enables assessing and comparing transferability across multiple source models with different architectures under domain shift, without requiring target labels and source data. Based on above, we introduce a Selection, Ensemble, and Adaptation (SEA) framework to address Zoo-MSFDA, which consists of: 1) source models selection based on the proposed principles and SUTE; 2) ensemble construction based on SUTE-estimated transferability; 3) target-domain adaptation of the ensemble model. Evaluations demonstrate that our SEA framework, with the introduced Zoo-MSFDA setting, significantly improves adaptation performance (e.g., 13.5\% on DomainNet). Additionally, our SUTE achieves state-of-the-art performance in transferability estimation.
\end{abstract}

}

\maketitle

\IEEEdisplaynontitleabstractindextext

%
\IEEEpeerreviewmaketitle

\input{introduction_qc}

\input{related_work}
\input{method_MMDA}

\input{method_selection}
\input{method_SEA}
\input{experiment}
\input{conclusions}


%

\ifCLASSOPTIONcaptionsoff
  \newpage
\fi



%
\bibliographystyle{IEEEtran}
\bibliography{mainbody}

\begin{IEEEbiographynophoto}{Jiangbo Pei} received the bachelor’s degree in communication engineering from the Beijing University of Posts and Telecommunications. He is working toward the PhD degree at the Beijing University of Posts and Telecommunications. Currently, he is also affiliated with the National Institute of Health Data Science, Peking University.
\end{IEEEbiographynophoto}

\begin{IEEEbiographynophoto}{Ruizhe Li} received the bachelor’s degree in Internet of Things from the Beijing University of Posts and Telecommunications. He is working toward the Master degree at the Beijing University of Posts and Telecommunications.
\end{IEEEbiographynophoto}

\begin{IEEEbiographynophoto}{Aidong Men} is a professor at the School of Artificial Intelligence, Beijing University of Posts and Telecommunications. His research interests include multimedia communication, digital TV, and images and speech signal processing and transmission. Men is a Fellow of the Chinese Institute of Electronics and China Institute of Communications. He is also an Invited Fellow of the Science and Technology Committee of State Administration of Radio, Film, and Television.
\end{IEEEbiographynophoto}

\begin{IEEEbiographynophoto}{Yang Liu} received PhD and MPhil in Advanced Computer Science from University of Cambridge, and B.Eng. in Telecommunication Engineering from Beijing University of Posts and Telecommunications (BUPT). She is now a Tenure-track Assistant Professor (Ph.D. Supervisor) in Wangxuan Institute of Computer Technology, Peking University.
\end{IEEEbiographynophoto}

\begin{IEEEbiographynophoto}{Xiahai Zhuang} is a professor at the School of Data Science, Fudan University. He graduated from Department of Computer Science, Tianjin University, received Master degree from Shanghai Jiao Tong University and Doctorate degree from University College London. His research interests include interpretable AI, medical image analysis and computer vision. His work won the Elsevier-MedIA 1st Prize and Medical Image Analysis MICCAI Best Paper Award 2023.
\end{IEEEbiographynophoto}

\begin{IEEEbiographynophoto}{Qingchao Chen} received the B.Sc. degree in telecommunication engineering from the Beijing University of Post and Telecommunication and the Ph.D. degree from the University College London. He was a Postdoctoral Researcher with University of Oxford, U.K (2018-2021). He is currently a assistant professor the National Institute of Health Data Science, Peking University. His current researches focus on computer vision and machine learning, radio-frequency signal processing and system design, and biomedical multimodality data analysis.
\end{IEEEbiographynophoto}

%






\end{document}

%% file: introduction_qc.tex
\section{Introduction}
\IEEEPARstart{T}{he} power of deep neural networks has been witnessed in various image classification tasks, yet their peak performance demands a substantial volume of high-quality data annotations. Consequently, there has been a notable emergence of interest in Unsupervised Domain Adaptation (UDA), which aims at transferring knowledge from a labeled source domain to an unlabeled target domain, overcoming the domain shift/discrepancy \cite{ganin2016domain,shen2021cdtd,long2015learning,ben2010theory,xu2014guest,xu2022delving}. However, traditional UDA algorithms raise concerns about personal data privacy and data transmission expenses, as they necessitate access to source data during adaptation.
To address this issue, Source-Free Domain Adaptation (SFDA) has been introduced as a promising alternative. SFDA focuses on learning a discriminative model for the unlabeled target domain by leveraging a model trained on the source domain without accessing the source data. However, the assumption of a single source domain in SFDA proves to be a limitation in the era of big data where diverse data acquisition resources are available \cite{dong2021confident}.

\begin{figure}
  \centering
\includegraphics[width=1.0\linewidth]{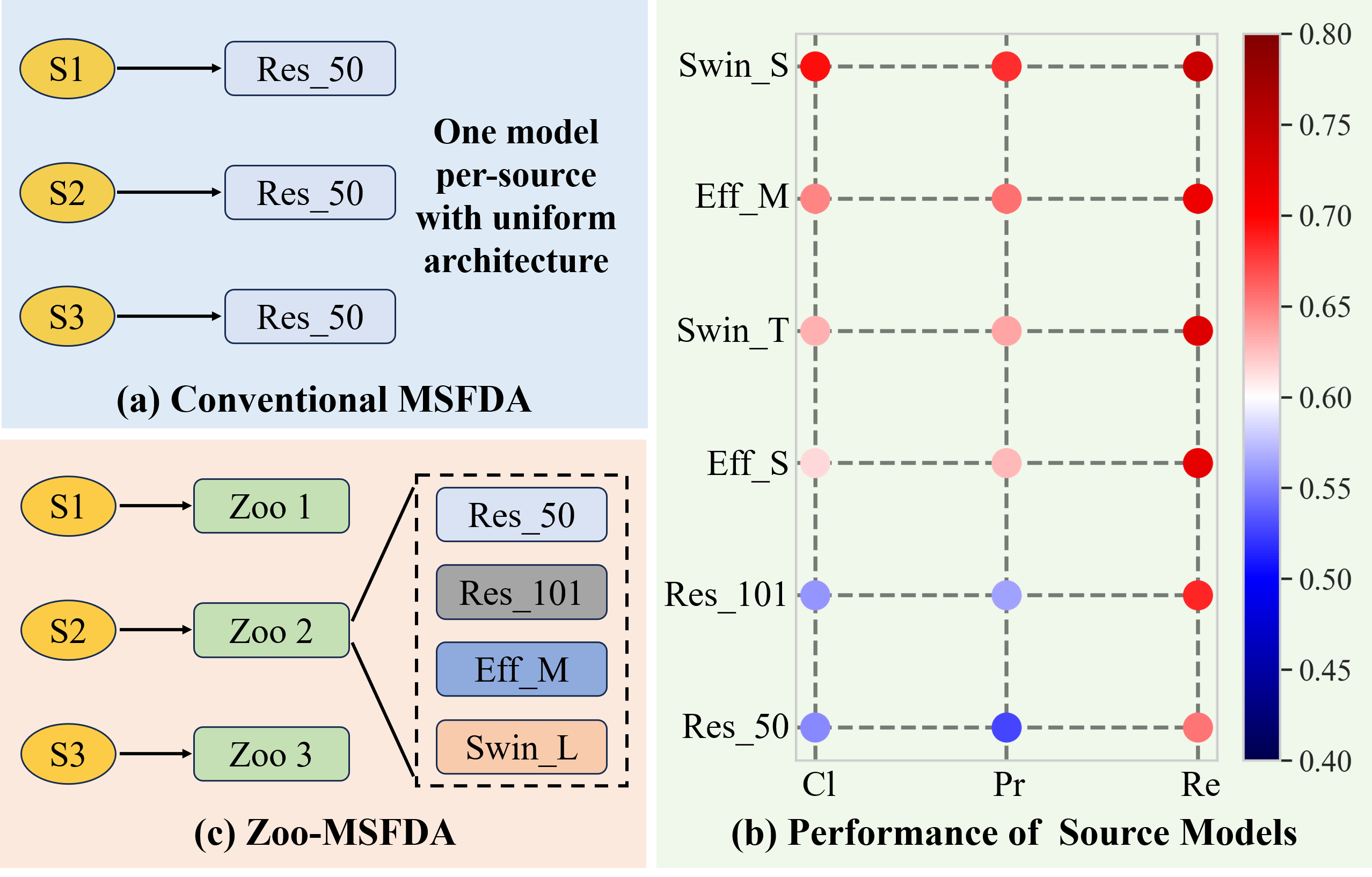}
\caption{(a) In previous MSFDA, each source domain (S1, S2, and S3) solely provides one source model, and all source models follow a uniform architecture. (b) Performance of source models on the target domain (evaluated on Office-Home \cite{venkateswara2017deep}: Cl, Pr, Re $\rightarrow$Ar). Each point represents a source model, where the color indicates its classification accuracy on the target domain. The horizontal axis represents the source domain from which the model originates. The vertical axis represents the model's architecture, i.e., ResNet50 \cite{he2016deep}, ResNet101 \cite{he2016deep}, EfficientNet\_V2\_S \cite{tan2021efficientnetv2}, EfficientNet\_V2\_M \cite{tan2021efficientnetv2}, Swin\_T \cite{liu2021swin}, and Swin\_S \cite{liu2021swin}. The results demonstrate that besides the source domain, the architecture of the source model also plays a pivotal role in influencing its performance on the target domain. 
(c) In our Zoo-MSFDA, we allow each source domain to offer a large zoo of source models with different architectures. The target user is permitted to access and leverage any model from these model zoos.}
   \label{fig:teaser}
\end{figure}

Recently, Multi-Source-Free Domain Adaptation (MSFDA) has garnered increasing attention as it maintains the protection of source data privacy while incorporating \textit{multiple source domains} into the framework \cite{ahmed2021unsupervised,pei2024evidential,dong2021confident,feng2021kd3a}. In MSFDA, it is typically assumed that each source domain solely provides a \textit{single} source model, and all models from multiple source domains follow a \textit{uniform} architecture, as shown in Fig. \ref{fig:teaser} (a). Although this conventional benchmark may aim for fair ablation comparisons, we argue that it is limited in the following aspects: 1) 
\textbf{Overlooking the effect of the source model architecture}: In addition to domain discrepancies, the architecture of source models also appears to be a crucial factor in adaptation (Fig. \ref{fig:teaser} (b)), as it is strongly related to the knowledge acquired by these models from their respective domains. However, this factor has not been thoroughly investigated in the conventional settings.
2) \textbf{Limited performance:} In our experiments, we observed \textit{the ground-breaking results} (e.g., 14.5\% improvement in the Office-Home \cite{venkateswara2017deep} dataset; Table \ref{tab_home_1}) if we select appropriate multi-architecture models from source domains and ensemble them in a straightforward method even \textit{without any adaptation modules}. Adhering strictly to uniform architectures may neglect more effective solutions that could leverage the diverse knowledge brought by multi-architecture source models.

The above analysis motivates us to explore the utilization of multi-architecture source models in MSFDA.
In this paper, we propose a new MSFDA setting named Zoo-MSFDA. As shown in Fig. \ref{fig:teaser} (c), we allow each source domain to offer \textit{a large zoo} of trained source models with different architectures. The target user is permitted to access and leverage any model from these model zoos, with the aim of learning a discriminative model for the unlabeled target domain.
Compared to MSFDA, Zoo-MSFDA provides the target user with a more comprehensive knowledge base from source domains, hence presenting greater potential for attaining optimal performance, and with unique challenges as well. 

\textit{Challenge 1: Missing analysis of source model selection principles.} One might argue that Zoo-MSFDA could be directly addressed by employing conventional MSFDA methods. However, this approach is impractical because the inclusion of numerous source models also increases the likelihood of including undesirable ones, which leads to significant performance degradation (Experiment \ref{sec:exp1}). 
This underscores the necessity for a principled model selection mechanism in Zoo-MSFDA to judiciously choose suitable source models while excluding unsuitable ones. However, previous model selection works (i.e., selecting models from model zoos) focus on either supervised scenarios \cite{mohr2023fast,you2021logme,bolya10495258scalable} or the vanilla UDA \cite{hu2024mixed,bachu2023building}, while the suitable selection principle in Zoo-MSFDA remains unexplored.

To address this challenge, we present a theoretical analysis for identifying appropriate source models in Zoo-MSFDA. This analysis unveils two fundamental selection principles. The first principle, termed as the \textit{transferability principle}, emphasizes selecting models that can accurately approximate the data distribution in the target domain. The second principle, termed as the \textit{diversity principle}, advocates for ensuring diversity and complementarity among the selected models to enhance the collective knowledge. We further introduce an algorithm that integrates both of them to implement source model selection in Zoo-MSFDA.

\textit{Challenge 2: Transferability estimation of models from different source domains with different architectures for the unlabeled target domain.} While measuring the diversity of source models can be readily accomplished using existing methods \cite{chen2023explore,bahng2020learning}, assessing their transferability poses a considerable challenge in Zoo-MSFDA. Most existing transferability measurements \cite{tran2019transferability, nguyen2020leep,you2021logme} require target labels, which are unavailable in Zoo-MSFDA. Conventional UDA methods \cite{long2015learning,ben2010theory} regard the transferability as the capacity to mitigate domain discrepancies. This approach requires access to source data, which is also inaccessible in Zoo-MSFDA. Pei \textit{et al.} \cite{pei2023uncertainty} posit a connection between uncertainty and transferability, advocating that models with lower uncertainties exhibit greater transferability. These methods are inefficient in Zoo-MSFDA due to the robustness issues in uncertainty measurements when confronted with distribution shifts \cite{ovadia2019can}. Recently, several target-only transferability estimations have emerged \cite{peng2024energy, yu2022predicting, hu2024mixed}. These methods enable the prediction of the transferability of source models to the target domain solely relying on unlabeled target data. These methods demonstrate effectiveness only when all source models conform to the same architecture. However, they suffer considerable performance degradation in Zoo-MSFDA where source models possess different architectures (Experiment \ref{sec:exp3}).

To tackle this challenge, we propose a novel \textbf{S}ource-Free \textbf{U}nsupervised \textbf{T}ransferability \textbf{E}stimation (SUTE). Aligned with UDA methods, we regard the transferability as the capacity of the model to overcome domain discrepancy. Our objective is to develop proxy indicators that could indirectly assess this capacity due to the unavailability of domain discrepancy in source-free scenarios. Through analysis, we identify three key indicators that are significantly associated with a model's transferability: \textit{individual certainty}, \textit{semantics consistency}, and \textit{global dispersity}. Based on them, we introduce the formulation of SUTE, which enables the assessment and comparison of transferability across multiple source models with \textit{different architectures} in the context of domain shift, \textit{without requiring access to any target labels or source data}. Remarkably, although with fewer requirements, SUTE achieves superior performance compared to existing transferability measurements.

To address the Zoo-MSFDA in general, we propose a new framework, namely Selection, Ensemble, and Adaptation (SEA). SEA consists of three steps: \textit{source model selection}, \textit{selected model ensemble}, and \textit{ensemble model adaptation}. In the \textbf{first} step, we utilize the proposed transferability principle (supported by our SUTE) and diversity principle to perform model selection on the received source models. 
This step ultimately selected a set of models that are identified appropriate for adaptation to the target domain, which are referred to as \textit{inlier models}. Other models are considered superfluous or risky and are denoted as \textit{outlier models}.
In the \textbf{second} step, we construct an ensemble model by combining the selected inlier models, aiming to effectively leverage the source knowledge they contain.
Different from previously done that based on learned transferability factors/domain weights \cite{dong2021confident,ahmed2021unsupervised}, our approach directly leverages the proposed SUTE for ensemble weighting due to its effectiveness in transferability estimation. 
In addition, considering that the outlier models may be valuable to certain target instance, we introduce an Outlier Knowledge Recycle module that guides the ensemble model to carefully identify and recycle useful knowledge from outlier models at the instance level.
In the \textbf{third} step, we introduce a Separate Information Maximization (SIM) objective to further adapt the ensemble model to the target domain. This objective is derived from previous (collaborative) Information Maximization \cite{dong2021confident,ahmed2021unsupervised}, with improvements in the utilization of diversity knowledge from source models.
We conduct extensive experiments to verify the effectiveness of our method and demonstrate that the proposed method achieves state-of-the-art results in terms of both \textit{adaptation performance} and \textit{transferability estimation}. 
Besides the exploitation of multi-architecture models, Zoo-MSFDA also encompasses sub-settings where models share the same architecture but are trained with multi-configurations such as learning rate, batch size, optimizer, and pre-trained weights (see Supplementary Fig. 3).
It is evident that our method is consistently effective in all sub-settings.

Our main contributions can be summarized as follows. 

\begin{itemize}
    \item 
    We introduce a new setting termed Zoo-MSFDA, which allows each source domain to offer multiple source models with different architectures. 
    Compared to the conventional MSFDA, it provides the target user with a more comprehensive knowledge base
    from source domains, presenting greater potential for
    attaining better adaptation performance.
    
    \item We delve into the model selection problem in Zoo-MSFDA. Based on theoretical analysis, we introduce two fundamental selection principles that guide
    effective source model selection, namely \textit{transferability principle} and \textit{diversity principle}.
    
    \item We propose a novel \textbf{S}ource-Free \textbf{U}nsupervised \textbf{T}ransferability \textbf{E}stimation (SUTE). It enables the assessment and comparison of transferability across multiple source models with \textit{different architectures} in the context of domain shift, \textit{without requiring access to any target labels or source data}. With fewer requirements, SUTE demonstrates superior performance compared to existing transferability measurements.
    
    \item We introduce a novel Selection, Ensemble, and Adaptation (SEA) framework that addresses Zoo-MSFDA by 1) selecting appropriate source models based on the proposed selection principles and SUTE; 2) constructing an ensemble model based on the proposed SUTE to  efficiently and safely aggregate source knowledge; and 3) adapting the ensemble model to the target domain using the proposed Separate Information Maximization (SIM). 
    
    \item We validate the effectiveness of our method through numerous experiments, and demonstrate that our approach achieves state-of-the-art results in both \textit{adaptation performance} and \textit{transferability estimation}. Besides the architecture, our method demonstrates consistent effectiveness in leveraging multiple source models trained with different learning rate, batch size, optimizer, and pre-trained weights.

\end{itemize}

%% file: related_work.tex
\section{Related Work}
\subsection{Source-Free Domain Adaptation}
Unsupervised Domain Adaptation (UDA) aims to leverage transferable knowledge from a source domain to enhance predictions in an unlabeled target domain. In the past few years, vanilla UDA approaches \cite{ganin2016domain, ben2010theory, gretton2006kernel} have achieved considerable success. However, UDA requires access to source data during the adaptation process, which may not be practical in various applications. Consequently, Source-Free Domain Adaptation (SFDA) has garnered increasing attention, which aims to adapt a source-trained model to an unlabeled target domain without accessing the source data. To address SFDA, recent studies attempt to generate surrogate source data \cite{ding2022source, tian2021source} and generate pseudo-labels for target data based on the target data structure \cite{liang2020we, ding2022source, liang2021source, zhang2022divide, yang2021exploiting}.
Although these methods have shown promising results, all SFDA approaches rely on the assumption that only one source domain is available, which limits their applicability in real-world settings where multiple source domains are often available.

\subsection{Multi-Source-Free Domain Adaptation}
Multi-Source-Free Domain Adaptation (MSFDA) maintains the protection of SFDA on the privacy of source data while incorporating diverse source domains into the framework \cite{ahmed2021unsupervised,pei2024evidential,dong2021confident,feng2021kd3a}.
Generally, MSFDA faces two key challenges: how to aggregate source knowledge and how to infer target semantics.
To address the first challenge, existing MSFDA methods propose to learn the domain weights\cite{ahmed2021unsupervised,feng2021kd3a} or transferable factors\cite{dong2021confident} to represent the contributions of source models to the \textit{entire} target domain, and then using them to combine source models to form an ensemble. Pei \textit{et al.} \cite{pei2024evidential} introduce a instance-level aggregation strategy based on evidential learning.
To tackle the second challenge, existing MSFDA methods mainly adopt the pseudo-label learning strategies. Ahmed \textit{et al.} \cite{ahmed2021unsupervised} and Feng \textit{et al.} \cite{feng2021kd3a} propose to generate pseudo labels by performing advanced clustering and majority voting, respectively.
Dong \textit{et al.} \cite{dong2021confident} introduce CAiDA, which leverages the local structure information of the target data to enhance the quality of the pseudo-labels.
While existing MSFDA methods have demonstrably advanced the field, they are limited by two key assumptions: 1) each source domain contributes a single model, and 2) all source models adhere to a uniform architecture. These assumptions restrict the investigation into the impact of source model architectures and potentially hinder the development of more effective adaptation methods that exploit multi-architecture source models.

\subsection{Transferability Measurements}
Assessing transferability of pre-trained models has great significance to guide common practice. Some methods focus on predicting the performance of the model in the conventional supervised tasks by using cross-validation, bootstrapping, or constructing learning curves \cite{mohr2022lcdb,mohr2023fast,viering2022shape}.
Recent works mainly focus on predicting the performance of a pre-trained source model in the target domain after fine-tuning in the supervised manner, where the target labels are available \cite{tran2019transferability,bolya2021scalable,nguyen2020leep,you2021logme}. For example, LEEP \cite{nguyen2020leep} introduces an empirical predictor based on estimating the joint distribution over pre-trained labels and the target labels, and uses the log expectation of the empirical predictor as a transferability measure. LogME \cite{you2021logme} formulates a logarithm of maximum evidence based on extracted features for efficient transferability estimation. Despite their success, the requirement of target labels in the above methods hinders their application in Zoo-MSFDA.

There are also some works focus on transferability estimation in UDA tasks \cite{sugiyama2007covariate,bachu2023building}. 
Conventional UDA works \cite{long2015learning,ben2010theory} regard the transferability as the capacity to mitigate domain discrepancies, which can be estimated by using domain discrepancy measurement such as Maximum Mean Discrepancy (MMD) \cite{gretton2006kernel} and A-Distance \cite{ben2010theory}.
Moreover, Sugiyama \textit{et al.} \cite{sugiyama2007covariate} introduce Importance-Weighted Cross-Validation (IWCV),
which estimates the target risk by re-weighting the source risk based on input-level domain similarity.
Based on it, You \textit{et al.} \cite{you2019towards} propose Deep Embedded Validation (DEV) that considers feature-level similarity and controls variance in IWCV. However, these methods are also impractical in Zoo-MSFDA due to their reliance on source data.

Recently, several target-only transferability estimations have been proposed for UDA tasks, which rely solely on unlabeled target data. Morerio \textit{et al.} \cite{morerio2017minimal} use the uncertainty (entropy) of the target predictions for validation, inspired by low-density assumption. Nonetheless, existing uncertainty measurements suffer from robustness issues in the presence of distribution shifts \cite{ovadia2019can}. Saito \textit{et al.} \cite{saito2021tune} introduce Soft Neighborhood Density (SND), which leverages neighborhood consistency for transferability measurement. Hu \textit{et al.} \cite{hu2024mixed}
propose MixVal, which utilizes the intra-cluster mixed samples for evaluating neighborhood density and the inter-cluster mixed samples for investigating the classification boundary. Although these methods do not require target labels and source data, their effectiveness is restricted to scenarios where all models share a common architecture. As we demonstrate in Experiment \ref{sec:exp3}, these methods suffer from performance degradation in the challenging Zoo-MSFDA where source models can have different architectures.

\subsection{Model Selection from Zoos}
Recently, selecting appropriate models from model zoos has gained increasing attention \cite{tran2019transferability}.
In prior studies, the model selection task has frequently been equated with the task of measuring transferability, where researchers opt for the most transferable model as their choice \cite{tran2019transferability,bolya2021scalable,nguyen2020leep,you2021logme,hu2024mixed}. Despite its simplicity, relying solely on a single model may not yield optimal results \cite{ahmed2021adaptive}. Consequently, the guiding principles for selecting models that lead to superior adaptation performance remain largely unexplored.
Notably, Agostinelli \textit{et al.} \cite{agostinelli2022transferability} recently introduce the pioneering work of extending transferability estimation to the selection of a set of source model ensembles, leveraging the concept of LEEP \cite{nguyen2020leep}. Building upon this foundation, Bachu \textit{et al.} \cite{bachu2023building} have furthered the field by incorporating considerations of domain mismatch within the latent feature representation space and exploring the interactions and correlations among model outputs.
However, these methods focus on scenarios of fine-tuning or the vanilla UDA, which are not suitable in Zoo-MSFDA.

Additionally, several Domain Generalization (DG) works have investigated leveraging model zoos during training on \textit{labeled source domains} \cite{chen2023explore,dong2022zood}.
Their primary goal is to enhance the generalizability of the source-trained model. In this paper, we explore the exploitation of model zoos on an \textit{unlabeled target domain} to enhance the adaptation on this domain, which is different from zoo-based DG works \cite{chen2023explore,dong2022zood} in both purpose and methodology.

%% file: method_MMDA.tex
\section{Zoo-MSFDA: Problem Definition and Theoretical Analysis}
\subsection{Problem Definition}
\label{sec:adapt}
In this paper, we consider a $C$-way classification task and assume the target domain shares the same labels with all source domains. Let $\{{D}_{S_k}\}_{k=1}^{K}$ denotes $K$ source domains, where ${D}_{S_k}$ contains $\vert {D}_{S_k}\vert $ labeled samples $\{(x_i,y_i)\}_{i=1}^{\vert {D}_{S_k}\vert }$. Each source domain ${D}_{S_k}$ provides a model zoo for the target domain, composed of $M^k$ source models trained on the labeled samples of this domain, i.e., $\{h_k^m\}_{m=1}^{M^k}$. $h_k^m(x_i)$ denotes the output probability of model $h_k^m$ for a given input $x_i$.
The target domain $D_T$ contains $\vert {D}_T\vert $ unlabeled samples, represented as $\{x_i\}_{i=1}^{\vert {D}_T\vert }$. The final goal of Zoo-MSFDA is to learn a target model $h_T$ based on these available source models and the unlabeled target data.  
All mathematical symbols used in this paper are summarized in Supplementary Table 1 for reader convenience.

\subsection{Theoretical Analysis}\label{sec:source_select}
In this section, we aim to provide theoretical insights for the model selection problem in the context of Zoo-MSFDA.
Without loss of generality, our analysis is conducted within the context of the pseudo-label learning paradigm, which has been widely employed in previous source-free tasks (e.g., the vanilla SFDA domain\cite{liang2020we,zhang2022divide}, MSFDA \cite{ahmed2021unsupervised,dong2021confident}, universal SFDA\cite{kundu2020universal}).

We begin by providing a brief introduction to the pseudo-label learning paradigm. Utilizing contemporary pseudo-label learning techniques (e.g., \cite{liang2020we,zhang2022divide}) and relying on a single source model $h_k^m$, we can derive the pseudo-label $\tilde{y}_i$ for the unlabeled target sample $x_i\in D_T$. This establishes a pseudo target distribution $P_{XY}^{h_k^m}$ that generates the data and pseudo-label pairs $D_{pse}^{h_k^m}=\{(x_i,\tilde{y}_i)\}_{i=1}^{\vert {D}_T\vert}$.
In the case of \textit{multiple} source models, current methods \cite{ahmed2021unsupervised,dong2021confident}
propose to aggregate the source models by ensemble approach (linear combination), utilizing a series of weights $\{ w_{k}^{m} \mid 1 \leq k \leq K, 1 \leq m \leq M^k \}$, satisfying the constraint $\sum_{k=1}^K \sum_{m=1}^{M^k} w_{k}^{m}=1$. Then, an ensemble model is obtained, which is defined as $\overline{h}(x)=\sum_{k=1}^K \sum_{m=1}^{M^k} w_{k}^{m} h_k^m(x)$.
Based on this, a new pseudo-target distribution is established: $P_{XY}^{\overline{h}}= \sum_{k=1}^K \sum_{m=1}^{M^k} w_{k}^{m} P_{XY}^{h_k^m}$. The corresponding data and pseudo-label pairs are denoted as $D_{pse}^{\overline{h}}=\{(x_i,\overline{y}_i)\}_{i=1}^{\vert {D}_T\vert}$. Then, given the loss function $\mathcal{l}: \mathcal{Y} \times \mathcal{Y} \rightarrow \mathbb{R}_{+}$, the overall objective (i.e., the empirical risk \cite{vapnik1991principles}) on the available unlabeled target data ${D}_T$ is formulated as:
 {\setlength\abovedisplayskip{1pt}
\setlength\belowdisplayskip{1pt}
\begin{equation}
\begin{aligned}
  \mathscr{R}_{emp}=\mathscr{L}(h_T,D_{pse}^{\overline{h}})=\frac{1}{^{\vert {D}_T\vert}}\sum_{i=1}^{\vert {D}_T\vert} \mathcal{l}(h_T(x_i),\overline{y}_i).
\end{aligned}
\label{eqa:pre_obj} 
\end{equation}}

Accordingly, the expected/true risk  on the real target domain distribution $P_{XY}^T$ can be represented by $\mathscr{R}_{exp}=\mathscr{L}(h_T,P_{XY}^T)$. The ultimate objective is to minimize the expected risk $\mathscr{R}_{exp}$ by optimizing the empirical risk $\mathscr{R}_{emp}$, which is known as empirical risk minimization \cite{vapnik1991principles}.
To achieve this objective, the key is to minimize the \textit{adaptation gap} between $\mathscr{R}_{exp}$ and $\mathscr{R}_{emp}$, which can be formulated by  $\mathscr{G}=\vert\mathscr{R}_{exp}-\mathscr{R}_{emp}\vert$.

\subsubsection{Transferability Principle}\label{sec:source_select_tra}

The following Proposition \ref{Proposition1} establishes an upper bound on the adaptation gap, facilitating analysis of its relationship to the involved source models, motivated by \cite{shen2023balancing}.

\begin{Proposition}\label{Proposition1} (proved in Supplementary 1.1)
Let $\mathscr{H}: X\rightarrow Y$ be a hypothesis space of multiclass predictors. Let $\mathscr{D}(.)$ be the Kullback-Leibler divergence.
Suppose $\mathscr{H}$ has finite Natarajan dimension $d(\mathscr{H})$, then for any loss function $\mathcal{l}$ bounded in $[0,z]$ and $h_T\in \mathscr{H}$, there exists a constant $\beta$ such that with a probability of $1-\delta$,
 {\setlength\abovedisplayskip{1pt}
\setlength\belowdisplayskip{1pt}
\begin{equation}
\begin{aligned}
  \mathscr{G} \leq\frac{\sqrt{2 z^2}}{2} \sqrt{\mathscr{D}(P_{XY}^{\overline{h}}\vert \vert P_{XY}^T)}
  +\beta  \sqrt{\frac{d(\mathscr{H}) \log C - \log \delta }{\vert D_T \vert}}.
\end{aligned}
\label{eqa:pro_1} 
\end{equation}}
\end{Proposition}
\noindent \textbf{\textit{Remark 1: Significance of Source Models' Transferability.}}
Proposition \ref{Proposition1} indicates that for a good selection principle, the selected models should enable a small discrepancy of  $\mathscr{D}(P_{XY}^{\overline{h}}\vert \vert P_{XY}^T)$, where $P_{XY}^T$ represents the true target distribution and $P_{XY}^{\overline{h}}$ is the pseudo-target distribution generated by $\overline{h}$, i.e., the ensemble of the selected source models.
We term this principle as the \textit{transferability principle}, as it underscores the ability of $\overline{h}$ in accurately capturing the target data distribution, i.e., the transferability \cite{saito2021tune} \footnote{Note that although we assume $\mathcal{l}$ is bounded, Proposition \ref{Proposition1} remains instructive for the general scenario where the unbounded cross-entropy is used. Please refer to Supplementary 4.4.}.

Despite the intuitive nature of the transferability principle, applying it effectively faces two significant challenges in real-world scenarios. The first challenge is \textit{how to measure the transferability of a model without accessing target labels}, which we will address in Section \ref{sec:trans}. The other challenge is \textit{how to apply the transferability principle efficiently}. Specifically, the number of combination strategies grows exponentially with the number of source models. Let $r$ be the total number of source models, the number of possible combinations involving inclusion or exclusion of each model is $2^r$ , resulting in  $2^r$ possible $\overline{h}$ even when ignoring the varying combination weights. Hence, it is time-consuming to measure the transferability for all possible ensembles $\overline{h}$.  
The subsequently introduced Equation \ref{eqa:coro_1} provides insights to address this challenge (proved in Supplementary 1.2). 

{\setlength\abovedisplayskip{1pt}
\setlength\belowdisplayskip{1pt}
\begin{equation}
\begin{aligned}
\mathscr{D}(P_{XY}^{\overline{h}}\vert \vert P_{XY}^T)\leq \sum_{k=1}^K \sum_{m=1}^{M^k} w_{k}^{m} \mathscr{D}(P_{XY}^{h_k^m}\vert \vert P_{XY}^T),
\end{aligned}
\label{eqa:coro_1} 
\end{equation}}where $w_{k}^{m}$ is a weight satisfying the constraint $\sum_{k=1}^K \sum_{m=1}^{M^k} w_{k}^{m}=1$, $\mathscr{D}(P_{XY}^{h_k^m}\vert \vert P_{XY}^T)$ measures the discrepancy between the true target distribution and that inferred by ${h_k^m}$, which represents the transferability of ${h_k^m}$ to the target domain.

\noindent \textbf{\textit{Remark 2: Priority of Leveraging Highly Transferable Individual Models.}} Equation \ref{eqa:coro_1} illustrates that the average transferability of individual source models serves as an upper bound to constrain the transferability of the ensemble model. Consequently, selecting source models with high transferability could serve as a valuable prior for ensembling. This motivates us to simplify the evaluation of the transferability by proposing a greedy strategy in Section \ref{sec:apply_tran}, which reduces the number of transferability evaluations from the exponential $2^r$ to $2r-1$, where $r$ denotes the total number of source models.

\subsubsection{Diversity Principle}\label{sec:source_select_div}
Besides the transferability, we introduce another perspective to analyse the Proposition \ref{Proposition1}.
The term $\mathscr{D}(P_{XY}^{\overline{h}}\vert \vert P_{XY}^T)$ measures the discrepancy between the true target distribution $P_{XY}^T$ and the pseudo-target distribution inferred by multiple source models $P_{XY}^{\overline{h}}=\sum_{k=1}^K \sum_{m=1}^{M^k} w_{k}^{m} P_{XY}^{h_k^m}$. To analyse this term, we introduce Proposition \ref{Corollary2} in the following.

\begin{Proposition}\label{Corollary2} (proved in Supplementary 1.3)
Let $\mathscr{L}(\boldsymbol{w})=\mathscr{D}(\sum_{k=1}^K \sum_{m=1}^{M^k} w_{k}^{m} P_{XY}^{h_k^m}\vert \vert P_{XY}^T)$.
Let $\boldsymbol{w^*}$ be the optimal weights, i.e.,  $\boldsymbol{w^*}=\arg \mathop{min}_{\boldsymbol{w}} \mathscr{L}(\boldsymbol{w})$, then, $\mathscr{L}(\boldsymbol{w^*})$ achieves minimum (0) if and only if $P_{XY}^T \in S_{\boldsymbol{w}}$, where 
$S_{\boldsymbol{w}}=\{\sum_{k=1}^K \sum_{m=1}^{M^k} w_{k}^{m} P_{XY}^{h_k^m} \mid \sum_{k=1}^K \sum_{m=1}^{M^k} w_{k}^{m}=1\}$. 
\end{Proposition}

\noindent \textbf{\textit{Remark 3: Significance of Source Models' Diversity.}} 
Proposition \ref{Corollary2} illustrates that, given the optimal ensemble weight $\boldsymbol{w^*}$, achieving the minimization of the discrepancy $\mathscr{D}(P_{XY}^{\overline{h}}\vert \vert P_{XY}^T)$ requires to satisfy the condition $P_{XY}^T \in S_{\boldsymbol{w}}$. This implies that enhancing the diversity among selected models to expand the set $S_{\boldsymbol{w}}$ is an efficient principle for reducing the adaptation gap. We term this principle as the \textit{diversity principle}. A toy example to understand the diversity principle is provided in Supplementary Fig. 2 (c).

\subsubsection{Trade-off between Transferability and Diversity}\label{sec:source_select_conf}
Although both transferability and diversity are crucial principles for source model selection, they can occasionally present trade-offs. For example, selecting models highly transferable to the target domain, which aligns with the transferability principle, might lead to a set with low diversity, where models share significant similarities (Supplementary Fig. 2 (b)).  From another perspective, prioritizing diversity may result in the incorporation of numerous poorly transferable models, which sacrifices the average individual transferability of the selected source models, thereby contradicting the argumentation in \textbf{\textit{Remark 2}}. 

In the exploration of a potential trade-off between the two principles, we observe that the transferability principle takes precedence over the diversity principle. This prioritization is reasoned by \textbf{\textit{Remark 3}}, which indicates that the diversity principle's effectiveness depends on achieving the optimal ensemble weight. However, as the target domain is unlabeled, the ability to attain the optimal ensemble weight is intricately linked to the initial transferability of the source models. Consequently, we prioritize the transferability principle over the diversity principle.


%% file: method_selection.tex
\section{Source Model Selection in Zoo-MSFDA}
\subsection{Overall}

\begin{figure*}
  \centering
\includegraphics[width=1.0\linewidth]{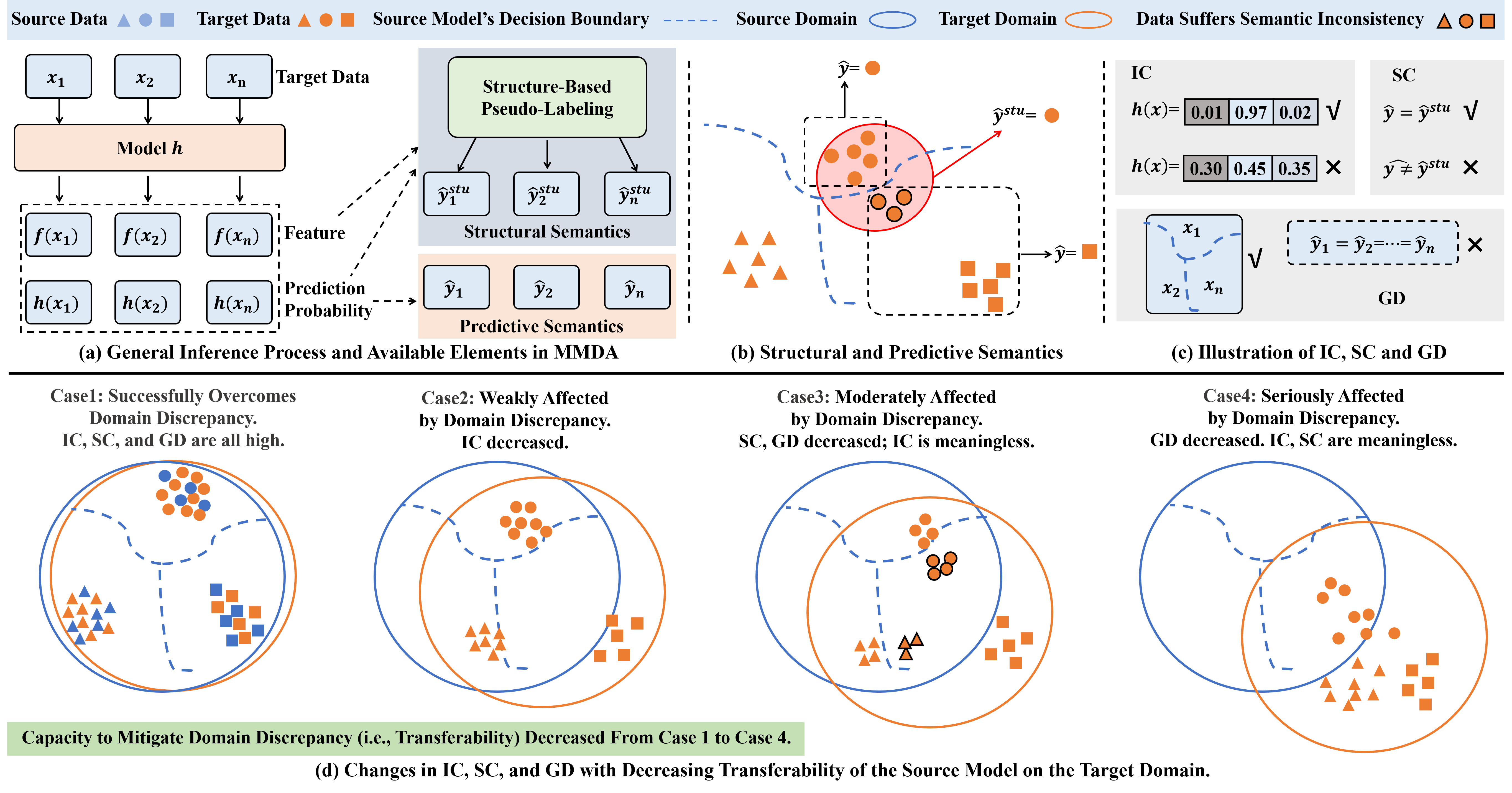}
    \caption{ (a) shows the general inference process of source models on target domain in Zoo-MSFDA. The extracted features, prediction probability, predictive semantics and structural semantics are elements available in the process. (b) illustrates the predictive semantics $\hat{y}$ and structural semantics $\hat{y}^{stu}$. The former is obtained based on the decision boundary, whereas the latter also considers the underlying data structure. (c) illustrates the three basic indicators of SUTE. \textbf{Individual Certainty}: A transferable source model is encouraged to provide a certainty prediction for each individual target sample. \textbf{Semantics Consistency:} Given a transferable source model, for a target sample, its predictive semantics (inferred from model prediction) and the structural semantics (inferred based on the cluster structure in the feature space), should be the same. \textbf{Global Dispersity:} Given a transferable source model, the distribution of predicted target abels is expected to exhibit diversity on the whole domain rather than collapsing into a few classes. Here we illustrate an extreme negative case of collapsing into a single classes, i.e., $\hat{y}^1=\dots=\hat{y}^n$. (d) shows a process of the model's transferability decrease (i.e., becoming more affected by domain discrepancy) and illustrates how IC, SC, and GD change during this process. A simulation demonstrating this process is provided in the Supplementary 2 and Supplementary Fig.1.}
   \label{fig:tran_illu}
\end{figure*}

In this section, we focus on tackling the source model selection problem in Zoo-MSFDA. Given multiple source models $H=\{ h_k^m \mid 1 \leq k \leq K, 1 \leq m \leq M^k \}$ from $K$ source domains, the goal is to select the appropriate source models (inlier models $S_I$) and filter out others models (outlier models $S_O$).

\textbf{Firstly}, while Section \ref{sec:source_select} has outlined two selection principles, the implementation of the transferability principles is challenging due to \textit{requisite but inaccessible target labels}, \textit{unaccessable source data} and \textit{architecture discrepancies}. 
To this end, we introduce a novel method termed Source-Free Unsupervised Transferability Estimation (SUTE) in Section \ref{sec:trans} (Fig. \ref{fig:tran_illu}). Here, three key indicators (individual certainty, semantics consistency, and global dispersity) associated with model transferability are formulated and integrated, effectively addressing the transferability estimation problem in Zoo-MSFDA.
\textbf{Subsequently}, we introduce the diversity measurement utilizing the Hilbert-Schmidt Independence Criterion (HSIC) in Section \ref{sec:diver}, allowing us to implement the diversity principle efficiently. \textbf{Finally}, we introduce a selection strategy that accomplishes model selection by incorporating the transferability and diversity principles.

\subsection{Transferability Estimation}\label{sec:trans}
While assessing the transferability of source models proves crucial, conducting a direct evaluation is impractical due to the need for target labels. Consequently, we approach the challenging task from an alternative perspective: evaluating models' transferability by estimating their capacity to mitigate domain discrepancies.
This view is closely related to vanilla UDA works \cite{long2015learning,ben2010theory}. However, the domain discrepancy measurements in these works necessitate access to source data, hindering their generalizability in Zoo-MSFDA.

In this paper, we propose a novel \textbf{S}ource-Free \textbf{U}nsupervised \textbf{T}ransferability \textbf{E}stimation (SUTE). This method enables the assessment and comparison of transferability amid domain shifts across multiple source models with the same or different architectures, without necessitating access to any source data or target labels. 

\subsubsection{Basic Elements}\label{sec:Basic_E}
Considering that comparisons of transferability may involve models with distinct architectures, we seek to leverage elements that generally exist across source models with different architectures. To this end, we illustrate the general inference process in Fig. \ref{fig:tran_illu} (a), and summary the following elements that are almost available for any source model $h$: 1) the extracted feature $f(x)$ for each target data $x$, 2) the output of the model  $h(x)$ for each target data $x$, representing the prediction probability $P_{y\vert x}$, 3) the predictive semantics (class) $\hat{y}=\arg min_{c} (h_c(x))$, where $h_c(x)$ represents the prediction probability on class $c$, and 4)
the structural semantics for each target data $x$, denoted as $\hat{y}^{stu}$, which could be obtained based on the extracted feature and the prediction probability by using some cluster-based pseudo-labeling technologies \cite{liang2020we,dong2021confident}. We illustrate the predictive semantics and the structural semantics in Fig. \ref{fig:tran_illu} (b).


\subsubsection{Transferability Indicator}\label{sec:Basic_C}

We address the challenging transferability estimation task by considering the question: \textit{Given a transferable source model that effectively overcomes the domain discrepancy, what characteristics should the above elements extracted from the model possess?}. The ideal scenario is that the distribution of target data should closely resemble that of the source data in the feature space. Based on this intuition, we summarize three necessary characteristics that a transferable source model should exhibit in the target domain, and propose three indicators namely Individual Certainty (IC), Semantics Consistency (SC) and Global Dispersity (GD). These indicators are illustrated in Fig. \ref{fig:tran_illu} (c).

\noindent \textbf{\textit{Individual Certainty.}}
Given a proficiently trained source model, the source data in the feature space should distinctly reside away from the model's decision boundary. 
For well-aligned target features, this characteristic should also hold. 
This means the transferable source model is encouraged to provide a certainty prediction for each target sample. We regard this characteristic as an indicator, and measure it using the negative entropy, formulated by: $\text{IC}=-\mathbb{E}_{x\sim D_T} \mathcal{H}(P_{y|x})=-\mathbb{E}_{x\sim D_T} \mathcal{H}(h(x))$.

\noindent \textbf{\textit{Semantics Consistency.}}
Besides the semantics from prediction (i.e., $\hat{y}$), the source data naturally exhibit structural semantics $\hat{y}^{stu}$ in the feature space of the source model. Generally, the semantics extracted from this structure align with those obtained from predictions. We term this characteristic as Semantics Consistency. This characteristic should persist for the target features if they align well with the source features. To this end, we quantify this characteristic by measuring the negative conditional entropy between $\hat{y}^{stu}$ and $\hat{y}$.  Then, the corresponding indicator SC can be formulated by: $\text{SC}=-\mathcal{H}(P_{\hat{Y}^{stu}\vert \hat{Y}})$, where $\hat{Y}^{stu}$ and $\hat{Y}$ represent the distribution of $\hat{y}^{stu}$ and $\hat{y}$, respectively.

\noindent \textbf{\textit{Global Dispersity.}}
Assuming the alignment of target features, the distribution of predicted target labels should generally exhibit diversity rather than collapsing into a few classes (Fig. \ref{fig:tran_illu} (c)).
We measure this characteristic using an indicator termed Global Dispersity, formulated it by $\text{GD}=\mathcal{H}(P_{\hat{Y}})=\mathcal{H}(\mathbb{E}_{x\sim D_T}P(\hat{y}\vert x))= \mathcal{H}(\mathbb{E}_{x\sim D_T}(h(x)))$.

Note that this formulation implicitly assumes a uniform label distribution of the target domain, since $\text{GD}$ achieved maximum when $P(y)=(\frac{1}{C}, \dots, \frac{1}{C})$. However, this assumption may not hold in practice. We will discuss and address this problem in the next Section. 

\subsubsection{SUTE Formulation}\label{sec:SUTE_Formulation}
We view the transferability of a source model to a target domain as its capacity to mitigate domain discrepancy among the two domains. Fig. \ref{fig:tran_illu} (d) visually illustrates the changes in IC, SC and GD across different levels of the transferability of a source model (we recommend referring to the simulation presented in \textbf{Supplementary 2} for better understanding).
\begin{itemize}
\item Initially, if the source model has robust transferability to the target domain that entirely overcomes domain discrepancies, as discussed in Section \ref{sec:Basic_C}, IC, SC, and GD are generally all high.
\item In scenarios where transferability weakens, the source model is weakly affected by domain discrepancies. As depicted in Fig. \ref{fig:tran_illu} (d), the model may still retain high prediction accuracy.  In such cases, SC and GD are not significantly influenced, while IC notably decreases.
\item When the source model is moderately affected by domain discrepancies, some samples may be misclassified, yet the structural information generally persists. This scenario is characterized by a significant decrease in SC.
\item If the source model is seriously affected by domain discrepancies, with common misclassifications and severe distortion of structural information, IC and SC become meaningless. This scenario can be characterized by a significant decrease in GD.

\end{itemize}

With the above consideration, we formulate the SUTE as:
 {\setlength\abovedisplayskip{1pt}
\setlength\belowdisplayskip{1pt}
\begin{equation}
\text{SUTE}= \lambda_1 \text{IC}+\lambda_2 \text{SC} + \Phi(\text{GD};\tau_h,\tau_l),
\label{eqa:SUTE} 
\end{equation}}where $\Phi(\text{GD};\tau_h,\tau_l)$ is a piecewise function of GD, formulated by:
{\setlength\abovedisplayskip{1pt}
\setlength\belowdisplayskip{1pt}
\begin{equation}
\Phi(\text{GD};\tau_h,\tau_l)=\left\{
\begin{aligned}
&\tau_h, & \text{GD}>\tau_h.\\
&\text{GD}, &\tau_l \leq \text{GD}\leq \tau_h.\\
&-\infty, & \text{GD}< \tau_l.\\
\end{aligned}
\right.
\label{eqa:clip} 
\end{equation}}

There are two reasons to use the piecewise function rather than the original GD. On the one hand, GD inherently favors a source model that yields an absolutely uniform label distribution in the target domain. However, our preference leans towards a source model that offers a relatively diverse prediction distribution, rather than the uniform label distribution, since it may not match the real target label distribution.
To this end, we set the  maximum value $\tau_h$. If GD reaches $\tau_h$, we consider it sufficiently diverse and do not require it to be larger. 

On the other hand, if the source model is seriously affected by domain discrepancy that leads to common misclassifications and severe distortion of structural information, IC and SC will be meaningless, as we illustrate in Fig. \ref{fig:tran_illu} (additional evidence is provided in Supplementary 2). Hence, when this case occurs (GD is smaller than $\tau_l$),
we directly set the SUTE to $-\infty$ to indicate the severely poor transferability, neglecting the influence of IC and SC.


\begin{figure*}
  \centering
 
  \includegraphics[width=0.85\linewidth]{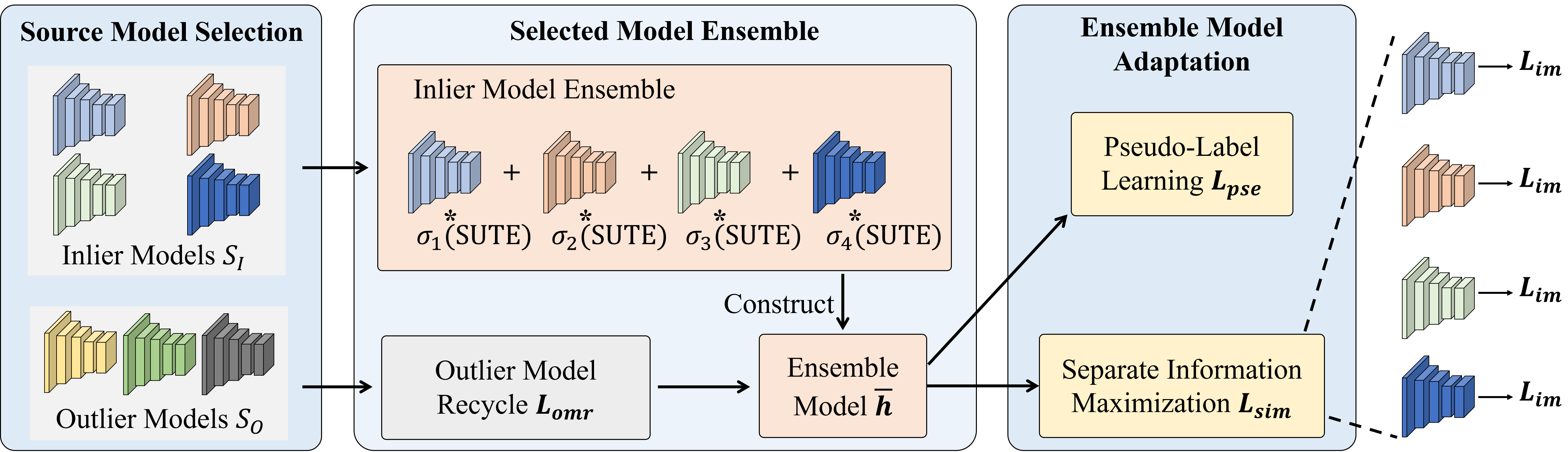}
\caption{Illustration of our SEA framework, which consists of three steps: (1) \textbf{Source Model Selection}. Given multiple source models, we obtain the inlier models $S_I$ and the outlier models $S_O$ based on the algorithms in Section \ref{sec:select}. (2) \textbf{Selected Model Ensemble}. In Inlier Model Ensemble, a series of SUTE-based weights are introduced to aggregate the inlier source models for constructing an ensemble model. Here, $\sigma$ represents the Softmax operator. 
In addition, an Outlier Model Recycle ($\mathcal{L}_{omr}$) is introduced to carefully recycle valuable knowledge from outlier models to the ensemble model.
(3) \textbf{Ensemble Model Adaptation}. Two objectives are introduced to adapt the ensemble model to the target domain. One is the widely-used pseudo-labeling objective $\mathcal{L}{pse}$. The other is the proposed Separate Information Maximization $\mathcal{L}{sim}$. It encourages information maximization for each inlier model separately, rather than for their entirety as in previous methods. After training, the ensemble model $\overline{h}$ will be employed as the target model $h_T$.}
\label{fig:method}
\end{figure*}

\subsection{Diversity Estimation}\label{sec:diver}
As highlighted in \textbf{\textit{Remark 3}} of Section \ref{sec:source_select_div}, enhancing the diversity of the selected source models is pivotal for improving the potential of ensemble. Over the past few years, numerous studies have proposed methods for selecting diverse models, primarily achieved by measuring the independence among these models. Inspired by \cite{bahng2020learning}, we adopt the Hilbert-Schmidt Independence Criterion (HSIC) for evaluating independence. Given two source models, denoted as $h_1$ and $h_2$, we represent HSIC by $\text{HSIC}(h_1, h_2)$. Detailed formulation of HSIC can be found in \cite{gretton2007kernel}.

\subsection{Source Model Selection}\label{sec:select}
Assume that there are multiple source models $H=\{ h_k^m \mid 1 \leq k \leq K, 1 \leq m \leq M^k \}$ from $K$ source domains, our aim is to perform source model selection based on the two principles. According to the discussion in Section \ref{sec:source_select_conf}, the transferability principle is considered a priority over the diversity principle. To this end, we firstly apply the transferability principle to select a set of transferable model, and then expand the diversity of the selected models to implement the diversity principle.

\subsubsection{Applying Transferability Principle}\label{sec:apply_tran}
We firstly collect a set of transferable models to form a transferable set $S_{TR}$.
Consider the SUTE of model $h_k^m$ (calculated by Equation \ref{eqa:SUTE}) is $\text{SUTE}_{h_k^m}$. We firstly create a model sequence $\text{Sort}(H)$, which is the sorted of source models in decreasing order of $\text{SUTE}_{h_k^m}$. 
Then, we sequentially add each model $h$ from $\text{Sort}(H)$ to the transferable set $S_{TR}$. Specifically, let $S_{TR}$ be the current transferable set, and after adding $h$, it becomes $S_{TR}'=S_{TR}\cup h$. Using ensemble strategy, we create two models: 1) the ensemble of $S_{TR}$, and 2) the ensemble of $S_{TR}'$. We regard each ensemble model as an individual model, and thus can calculate the SUTE for the two ensemble models (details are in Supplementary 3.3), respectively, denoted as $\text{SUTE}_{S_{TR}}$ and $\text{SUTE}_{S_{TR}'}$.
We then compare the two values, and only accept $h$ in $S_{TR}$ if the $\text{SUTE}_{S_{TR} \cup h}$ is larger than $\text{SUTE}_{S_{TR}}$. This means
the transferability of $S_{TR} \cup h$ is larger than that of $S_{TR}$, indicating that the transferability is improved after adding $h$. 
This procedure also ensures that the transferability of the ensemble of $S_{TR}$ can be no worse than the best individual model. In addition, this greedy strategy removes the need for manually specifying the number of selected models. 
The procedure is summarized in \textbf{Supplementary Algorithm 1.}

\subsubsection{Applying Diversity Principle} 
We then select a set of models to expand the diversity of the selected models. Specifically, 
we denote models that have not been selected ($h\notin S_{TR}$) as diversity candidate set $S_{DC}$. We calculate the average independence of each model $h\in S_{DC}$ to all models $h'\in S_{TR}$, represented by $\text{Div}(h)=\mathbb{E}_{h'\in S_{TR}} \text{HSIC}(h,h')$. We sort these models according to $\text{Div}(h)$ and selected the $\text{Top}_q$ models as Diversity set, i.e., $S_{DIV}=\{h\vert \text{Div}(h)\in \text{Top}_q(\text{Div}(h))\}$, where $q$ is a hyperparameter to control the number of $S_{DIV}$. 
The procedure is summarized in \textbf{Supplementary Algorithm 2.}

Finally, we add $S_{TR}$ and $S_{DIV}$ to obtain the inlier models $S_I = S_{TR} \cup S_{DIV}$. 
These models will be used to construct the target model and undergo adaptation. Other models that do not belong to $S_I$ are denoted as the outlier models, formulated by $S_O= \{h\vert h\in H, h\notin S_{I}\}$. These outlier models will neither be combined into target model nor be optimized during adaptation, which effectively avoids undesirable source models and reduces computational cost. 

%% file: method_SEA.tex
\section{Selection, Ensemble and Adaptation Framework}
\subsection{Overall}
We propose a novel Selection, Ensemble, and Adaptation (SEA) framework to address Zoo-MSFDA (Fig. \ref{fig:method}). SEA consists of three steps: \textit{source model selection}, \textit{selected model ensemble}, and \textit{ensemble model adaptation}.
Specifically, in the first step, we split source models to the inlier models $S_I$ and the outlier models $S_O$ based on the model selection algorithm in Section \ref{sec:select}.

In the second step, we propose an Inlier Model Ensemble, which aims to aggregate valuable knowledge from inlier models by ensembling these models. Different from previous MSFDA works, we directly use our SUTE for ensemble rather than using the learned domain weights. The ensemble model is formulated by $\overline{h}=\sum_{j=1}^{\vert S_I\vert} \sigma_{j}(\text{SUTE})\cdot h_{j}$. Additionally, we propose an Outlier Model Recycle objective ($\mathscr{L}_{omr}$) to meticulously recycle valuable knowledge from outlier models for further enhancing the ensemble model.

In the third step, we present a Separate Information Maximization objective $\mathscr{L}_{sim}$, an enhanced version of Information Maximization in MSFDA works, which leads to better adaptation performance in Zoo-MSFDA. The widely-used  pseudo-label learning strategy is also adopted, represented by $\mathscr{L}_{pse}$. The overall loss is represented by: $ \mathscr{L}_{all}= \mathscr{L}_{sim}+ \gamma_1 \mathscr{L}_{pse}+ \gamma_2 \mathscr{L}_{omr}$, where $\gamma_1$ and $\gamma_2$ are coefficients of $\mathscr{L}_{pse}$ and $\mathscr{L}_{omr}$, respectively.
After training, the ensemble model $\overline{h}$ will be employed as the target model $h_T$.

\subsection{Source Model Selection}
We follow the proposed procedure in Section \ref{sec:select} for source model selection, obtaining the inlier models $S_I$ and the outlier models $S_O$.

\subsection{Selected Model Ensemble}

\subsubsection{Inlier Model Ensemble}\label{sec:aggregation}

The inlier models $S_I$ selected through our meticulous design selection procedure contain substantial and safe-to-utilize source knowledge.
We construct an ensemble of inlier models to leverage the valuable source knowledge they contained sufficiently, following previous MSFDA methods \cite{dong2021confident,ahmed2021unsupervised}.
Specifically, these methods first formulate the weights for ensemble by: $\boldsymbol{\theta}=\{\theta_{j} \vert 1 \leq j \leq \vert S_I\vert\}$, satisfying the constraint $\sum_{j=1}^{\vert S_I\vert}  \theta_{j}=1$, where $\vert S_I\vert$ is the number of inlier models. Then, the ensemble model is formulated by $\overline{h}=\sum_{j=1}^{\vert S_I\vert}  \theta_{j} h_{j}$.
Intuitively, for the given source models $\{h_1, h_2, \dots, h_{\vert S_I\vert}\}$, if the optimal weights $\boldsymbol{\theta^*}:=\mathop{min}\limits_{{\boldsymbol{\theta}}}\mathscr{L}(\sum_{j=1}^{\vert S_I\vert} \theta_{j} h_{j})$ are achieved ($\mathscr{L}$ be any optimization objective), it is better than any single model:

{\setlength\abovedisplayskip{1pt}
\setlength\belowdisplayskip{1pt}
\begin{equation}
\begin{aligned}
\mathscr{L}(\sum_{j=1}^{\vert S_I\vert}  \theta^*_{j} h_{j})=\mathop{min}\limits_{{\boldsymbol{\theta}}}\mathscr{L}(\sum_{j=1}^{\vert S_I\vert}  \theta_{j} h_{j})\leq \mathop{min}\limits_{h_1, h_2, \dots, h_{\vert S_I\vert}}\mathscr{L}(h_{j}).
\end{aligned}
\label{eqa:agg_i} 
\end{equation}}

In  previous MSFDA methods \cite{dong2021confident,ahmed2021unsupervised}, the weights are generally obtained by learning. However, as the target domain is unlabeled, the attainment of the optimal weights could not be granted. Motivated by the evidence that our proposed SUTE could accurately infer models' transferability, instead of using learning-based methods, we propose to directly use the SUTE of these models to weight them. Hence, we formulate the weights by $\sigma(\text{SUTE})=\sigma([\text{SUTE}_1,\dots,\text{SUTE}_{\vert S_I\vert}])$, where $\sigma$ is the softmax operator for normalization, $\text{SUTE}_i$ denotes the SUTE of the $i$-th source model obtained by using Equation \ref{eqa:SUTE}. The ensemble model can be denoted as $ \overline{h}=\sum_{j=1}^{\vert S_I\vert}  \sigma_{j}(\text{SUTE})\cdot h_{j}$, where $\sigma_{j}(\text{SUTE})$ represents the $j$-th elements of $\sigma(\text{SUTE})$.

\subsubsection{Outlier Model Recycle}
Outlier models, despite their potential lack of substantial impact on overall adaptation or even the risk of negative effects, may still harbor knowledge valuable to specific target sample. To this end, we propose Outlier Model Recycle, a module for carefully leveraging the knowledge gleaned from these outlier models at the instance level. Given an instance $x$ and $h \in S_O$, we recycle the prediction $h(x)$ only when two strict constraints are satisfied. First, the confidence of $h$ in predicting $x$, i.e., the maximum probability $\max_c h_c(x)$, should be the highest among all outlier models. Let ${h}_c(x)$ be the prediction probability of model $h$ for input $x$ on class $c$, this constraint can be represented by: $\max_c h_c(x) = \max_{h' \in S_O} \max_c h_c'(x)$. Second, the confidence should exceed a threshold $\tau$. This constraint can be formulated by $\max_c h_c(x) > \tau$. If a prediction satisfies the above constraints, we store the data and the label inferred from this outlier 
model's prediction  (formulated by $y_{o} = \arg \max_c h_c(x)$) as a pair $(x, y_{o})$. Finally, let $D_{o}$ be the set of all collected pairs, the objective of Outlier Model Recycle $\mathcal{L}_{omr}$ can be formulated by:


{\setlength\abovedisplayskip{1pt}
\setlength\belowdisplayskip{1pt}
\begin{equation}
\begin{aligned}
\mathscr{L}_{omr}=\mathbb{E}_{(x,y_{o})\sim D_{o}} \mathcal{l}_{ce}(\overline{h}(x),y_o),
\end{aligned}
\label{eqa:pse} 
\end{equation}}where $\mathcal{l}_{ce}$ denotes the cross-entropy loss function.

\subsection{Ensemble Model Adaptation}\label{sec:adaptation}
\subsubsection{Pseudo-Label Learning}

Pseudo-label learning is a widely used adaptation method in SFDA/MSFDA works \cite{liang2020we,ahmed2021unsupervised,dong2021confident,yang2021exploiting}.
In particular, leveraging the pseudo labels derived from data structure in the feature space is a popular and effective method for adaptation.  However, we find that these methods lose efficiency in Zoo-MSFDA benchmarks where model architectures are different.

In our SEA, we instead use a simple and efficient way that directly using the predictive semantic of each sample as its pseudo label, which can be formulated by $\overline{y} = \arg \max_c \overline{h}_c(x)$, where $\overline{h}_c(x)$ represents the prediction probability of $\overline{h}$ on class $c$ given $x$. We represent these data and pseudo-label pairs as $D_{pse}^{\overline{h}}=\{(x_i,\overline{y}_i)\}_{i=1}^{\vert {D}_T\vert}$. Then, the objective of the pseudo-label learning
can be formulated by

{\setlength\abovedisplayskip{1pt}
\setlength\belowdisplayskip{1pt}
\begin{equation}
\begin{aligned}
\mathscr{L}_{pse}=\mathbb{E}_{(x,\overline{y})\sim D_{pse}^{\overline{h}}} \mathcal{l}_{ce}(\overline{h}(x),\overline{y}),
\end{aligned}
\label{eqa:pse} 
\end{equation}}where $\mathcal{l}_{ce}$ denotes the cross-entropy loss function.

\subsubsection{Separate Information Maximization}
Information Maximization \cite{liang2020we} has proved effectiveness  in adapting source models to the target domain \cite{dong2021confident,ahmed2021unsupervised,yang2022attracting}. This can be formulated as:
{\setlength\abovedisplayskip{1pt}
\setlength\belowdisplayskip{1pt}
\begin{equation}
\begin{aligned}
    \mathscr{L}_{im}(h,D_T) &= -\mathbb{E}_{{x \in D_T}}\mathcal{H}(h(x)) + \mathcal{H}(\mathbb{E}_{{x \in D_T}}(h(x))),
\end{aligned}
\end{equation}}where the first term aims at minimizing the entropy on model prediction, while the latter encourages the predicted empirical label distribution to be a uniform distribution.

\noindent\textbf{\textit{Information Maximization in Previous MSFDA Works.}}
Previous MSFDA works \cite{dong2021confident,ahmed2021unsupervised} directly apply the Information Maximization (IM) objective on the ensemble model. We term this approach Collaborative Information Maximization $\mathscr{L}_{cim}$.  Specifically, this can be formulated as:

{\setlength\abovedisplayskip{1pt}
\setlength\belowdisplayskip{1pt}
\begin{equation}
\begin{aligned}
    \mathscr{L}_{cim} =\mathscr{L}_{im}(\overline{h},D_T)=\mathscr{L}_{im}(\sum_{j=1}^{\vert S_I\vert}  \theta_{j} h_{j}),
\end{aligned}
\end{equation}}where $\overline{h}=\sum_{j=1}^{\vert S_I\vert} \theta_{j} h_{j}$ is the prediction probability of the target model. However, we observed that $\mathscr{L}_{pre\_im}$ may encourage the source models within the ensemble model to make the same prediction. This may hinder the mining of the diversity knowledge from source models.

\noindent\textbf{\textit{Our Separate Information Maximization.}}
We propose to independently optimize the IM loss for each model within the ensemble model, rather than for the whole ensemble model.
This decreases the identical-prediction constraint among source models, enhancing to activate and leverage diversity knowledge they contained.
The new objective can be formulated as follows:
{\setlength\abovedisplayskip{1pt}
\setlength\belowdisplayskip{1pt}
\begin{equation}
\begin{aligned}
   \mathscr{L}_{sim} =\sum_{j=1}^{\vert S_I\vert}  \theta_{j} \cdot \mathscr{L}_{im}(h_{j},D_T),
\end{aligned}
\end{equation}}where $\theta_{j}$ denotes the combination weight for $j$-th source model. We set $\theta_{j}=\sigma_{j}(\text{SUTE})$ for consistency with Section \ref{sec:aggregation}.

\subsection{Overall Objective}
\noindent\textbf{\textit{Overall Loss Function.}}
The final objective is the combination of all aforementioned loss functions, formulated by: {\setlength\abovedisplayskip{1pt}
\setlength\belowdisplayskip{1pt}
\begin{equation}
\begin{aligned}
  \mathscr{L}_{all}= \mathscr{L}_{sim}+ \gamma_1 \mathscr{L}_{pse}+ \gamma_2 \mathscr{L}_{omr},
\end{aligned}
\label{eqa:opt} 
\end{equation}}where $\gamma_1$ and $\gamma_2$ are the weights of $\mathscr{L}_{pse}$ and $\mathscr{L}_{omr}$, respectively. 

\noindent\textbf{\textit{Optimized Parameters.}}
Previous SFDA and MSFDA methods typically optimize the feature encoder of models while keeping the classifier fixed.  
In contrast, our approach solely optimizes the lightweight classifiers, while maintaining all feature encoders in a fixed state. This design choice capitalizes on the fact that, based on the rich source knowledge base provided by Zoo-MSFDA and our effective source model selection method, the feature encoders of the inlier models have generally contained sufficient and comprehensive knowledge. As a result, further optimization of the feature encoders is rendered unnecessary (see Supplementary 4.3; Supplementary Table 15). The final optimization objective can be represented by:
{\setlength\abovedisplayskip{1pt}
\setlength\belowdisplayskip{1pt}
\begin{equation}
\begin{aligned}
  \min_{g_1, g_2, \dots, g_{\vert S_I\vert}} \mathscr{L}_{all},
\end{aligned}
\label{eqa:opt2} 
\end{equation}}where $g_{j}$ represents the parameters in the classifier of $j$-th inlier model $h_j$.
After adaptation, the ensemble model is finally employed as the target model, i.e., $h_T=\overline{h}$.

%% file: experiment.tex
\section{Experiments}

\subsection{Experimental Setup}
\subsubsection{Dataset}
In the evaluation, we consider three common and challenging datasets: Office-Home \cite{venkateswara2017deep}, Office-31\cite{saenko2010adapting}, and DomainNet \cite{peng2019moment}.
Office-Home consists of four domains: Artistic images (A), Clipart (C), Product images (P), and Real-World images (R), featuring a total of 65 classes and 15,500 images. Following \cite{ahmed2021unsupervised,dong2021confident}, we set 4 transfer tasks on the Office-Home dataset, namely 1) A, C, P$\rightarrow$R, 2) A, C, R$\rightarrow$P, 3) A, P, R$\rightarrow$C, and 4) C, P, R$\rightarrow$A.
Office-31 contains 4,652 images in 31 categories from three domains: Amazon (A), Webcam (W) and DSLR (D). We conduct 3 transfer tasks on the Office-31 dataset: 1) A, W$\rightarrow$D, A, D$\rightarrow$W, and  D, W$\rightarrow$A.
DomainNet comprises approximately 0.6 million images categorized into 345 classes across six domains: Quickdraw (Q), Clipart (C), Painting (P), Infograph (I), Sketch (S), and Real (R). We conduct 6 transfer tasks on the DomainNet dataset: 1) I, P, Q, R, S$\rightarrow$C, 2) C, P, Q, R, S$\rightarrow$I, 3) C, I, Q, R, S$\rightarrow$P, 4) C, I, P, R, S$\rightarrow$Q, 5) C, I, P, Q, S$\rightarrow$R, and 6) C, I, P, Q,  R$\rightarrow$S.

\subsubsection{Networks and Hyperparameters}
For source training, detailed information regarding the number/architecture/performance of source models in different settings/datasets can be found in the Supplementary 3.1.
For target adaptation, all hyper-parameters of our SUTE/SEA either remain consistent across all datasets and settings, or are set as adaptive values. Details are in Supplementary 3.2.

\subsubsection{Evaluation Protocols}
We evaluate the proposed method through the following experimental studies.

\noindent \textbf{\textit{Experiment\#1: MSFDA vs. Zoo-MSFDA.}}
This experiment aims to compare the performance of existing methods in both the previous MSFDA setting and our Zoo-MSFDA setting. The following state-of-the-art MSFDA methods are considered: DECISION \cite{ahmed2021unsupervised},  CAiDA \cite{dong2021confident}, and KD3A \cite{feng2021kd3a}. Results in the MSFDA setting are reported from \cite{dong2021confident,feng2021kd3a}. 
Results in the Zoo-MSFDA setting are reproduced by ourselves.  

\noindent \textbf{\textit{Experiment\#2: Comparison with State-of-the-Art Methods.}}
This experiment compares our method with existing methods in the Zoo-MSFDA setting. We firstly apply state-of-the-art MSFDA methods, including DECISION \cite{ahmed2021unsupervised},  CAiDA \cite{dong2021confident}, and KD3A \cite{feng2021kd3a}.
Subsequently, we utilize some existing transferability estimating methods, thus allowing these MSFDA methods to perform source model selection before adaptation. The transferability estimating methods include  Average Negative Entropy (ANE) \cite{morerio2017minimal}, Negative Mutual Information (NMI) \cite{huang2022frustratingly,liang2020we}, Meta-Distribution Energy (MDE) \cite{peng2024energy}, LEEP \cite{nguyen2020leep} and LogME \cite{you2021logme}.
Note that to implement LEEP \cite{nguyen2020leep} and LogME \cite{you2021logme} in Zoo-MSFDA, the requisite target labels are substituted with pseudo labels obtained via the algorithm proposed by \cite{ahmed2021unsupervised}. We represent the new implementations of the two methods by LEEP$^*$ and LogME$^*$, respectively. Except MDE \cite{peng2024energy}, all transferability estimating methods follow the same procedure in Supplementary Algorithm 1 to select source models (by changing the SUTE to these methods). Given the invalidity of MDE when using Supplementary Algorithm 1, we chose to directly select the Top$_5$ models with the highest MDE values.

\noindent \textbf{\textit{Experiment\#3: Transferability Estimation Analysis.}}
This experiment compares the transferability estimation capabilities of the proposed SUTE with existing methods.
In addition to the measurements presented in Experiment\#2 (ANE, NMI, MDE, LEEP$^*$, and LogME$^*$), we also consider SND\cite{saito2021tune} and MixVal\cite{hu2024mixed}, which implicitly assume uniform architectures across source models.
Furthermore, we consider oracle methods requiring target labels or source data, including: standard LEEP (requiring target labels), standard LogME (requiring target labels), Maximum Mean Discrepancy (MMD) \cite{gretton2006kernel} (requiring source data), and A-Distance \cite{ben2010theory} (requiring source data). The evaluation metric is the Spearman's Rank Correlation Coefficient  \cite{spearman1987proof} between 1) each source model's performance on the target domain and 2) the estimated transferability of the model on the target domain. A larger Spearman's Rank Correlation Coefficient  represents superior transferability estimation ability.

\noindent \textbf{\textit{Experiment\#4: Ablation Analysis.}} In this experiment, we conduct ablation studies to evaluate the effectiveness of various components involved in our method.

\noindent \textbf{\textit{Experiment\#5: Analysis on Sub-Settings.}} In this experiment, we evaluate our method on 4 sub-settings (Zoo-MSFDA-S1/S2/S3/S4) to analyze the reliability of our method for other factors (expect the architectures) that lead to diversity of source models, including the \textit{learning rate}, \textit{batch size}, \textit{optimizer} and \textit{pre-trained weight}. The detailed implementation of these settings can be found in Supplementary 3.1.

\input{table/tab_home_1}
\input{table/tab_31_1}
\input{table/tab_dn_1}

\input{table/tab_home_2}

\input{table/tab_31_2}

\input{table/tab_dn_2}
\input{table/tab_home_tran}

\input{table/tab_31_tran}

\input{table/tab_dn_tran}

\input{table/tab_home_tran_ora}

\begin{figure*}
  \centering
 
  \includegraphics[width=0.95\linewidth]{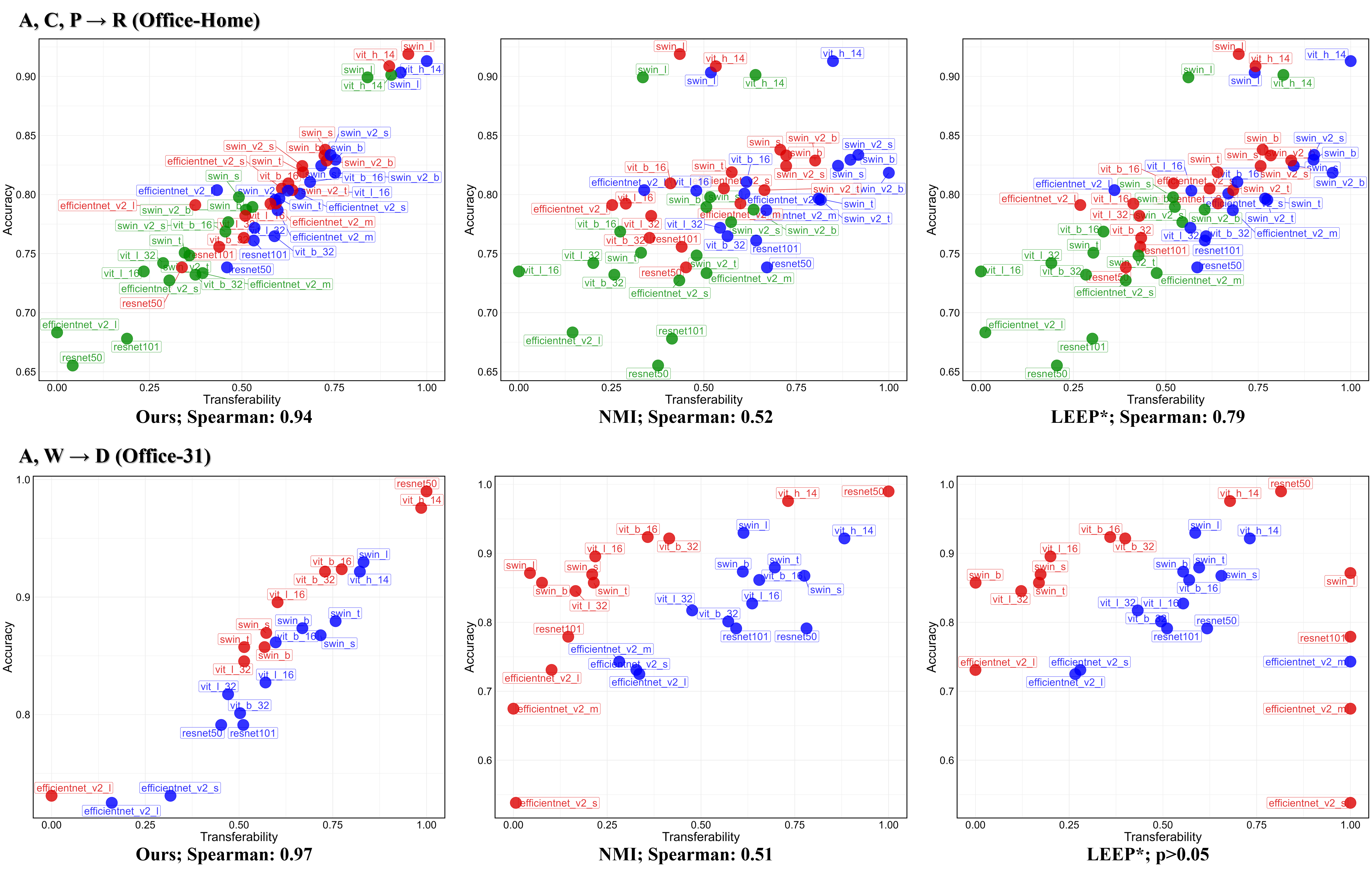}
\caption{ The joint distribution (in the Zoo-MSFDA setting) of 1) the performance of source models on target domain, and 2) the transferability of source models to the target domain estimated by using different methods. The Spearman's Rank Correlation Coefficient is used as the evaluation metric. The target domain is Office-Home's Re. The architecture of the source models has been annotated in the figure. \textbf{Top:} The joint distribution on Office-Home's task A, C, P$\rightarrow$R, where models from A, C, P are denoted by blue, green and red, respectively. \textbf{Bottom:} The joint distribution on Office-31's task A, W$\rightarrow$D, where models from A, W are denoted by blue and red, respectively. $p$-value $>$0.05 indicates the results do not have statistics significance.
}
   \label{fig:joint_Zoo-MSFDA}
\end{figure*}

\subsection{Experiment\#1: MSFDA vs. Zoo-MSFDA}\label{sec:exp1}

Table \ref{tab_home_1} and Table \ref{tab_dn_1} reports the performance of DECISION, CAiDA, and KD3A, in both the MSFDA and Zoo-MSFDA settings. We have the following observations.

\noindent\textbf{\textit{Results on the Office-Home Dataset.}} Since Zoo-MSFDA provided  more available source models, the performance  in the Zoo-MSFDA setting is generally superior compared to that in the previous MSFDA setting (Table \ref{tab_home_1}). Specifically, DECISION exhibits a performance improvement of 6.1\%. CAiDA demonstrates a performance increase of 4.9\%. KD3A outperforms its performance in the previous MSFDA setting by 6.9\%.

\noindent\textbf{\textit{Results on the Office-31 Dataset.}}
The performance of all methods in the Zoo-MSFDA setting surpasses that in the previous MSFDA setting (Table \ref{tab_31_1}). The improvements of DECISION, CAiDA, and KD3A are 3.1\%, 1.8\%, and 1.0\%, respectively.

\noindent\textbf{\textit{Results on the DomainNet Dataset.}}
As seen in Table \ref{tab_dn_1},existing methods in the Zoo-MSFDA setting suffer notable performance deterioration compared with their performance in the MSFDA setting\footnote{We were unable to implement CAiDA on DomainNet in the Zoo-MSFDA setting due to the computational cost.}. For example, DECISION decreased by 7.2\%, and KD3A decreased by 10.0\%. This demonstrates that \textbf{having more source models does not always mean better adaptation}, since it may also incorporate undesirable models.

\input{table/tab_other_tran}

\subsection{Experiment\#2: Comparison with State-of-the-Art Methods  in Zoo-MSFDA.}\label{sec:exp2}
To demonstrate the effectiveness of our method in addressing the Zoo-MSFDA, we conduct experiments to compare our method with existing MSFDA methods and their variants by combining them with existing model selection methods.

\noindent \textbf{\textit{Comparison with MSFDA Methods.}} As shown in Table \ref{tab_home_1}, Table \ref{tab_31_1} and Table \ref{tab_dn_1}, our method outperforms the state-of-the-art MSFDA methods by 9.7\% and 0.7\% in terms of average accuracy on the Office-Home and Office-31 datasets. More notably, our method brings significant improvement on the DomainNet dataset, surpassing the second-best one by 23.5\%.

\noindent \textbf{\textit{Comparison with Model-Selection-Based MSFDA Methods.}}
Model-Selection-Based MSFDA (MSB-MSFDA) methods are variants of existing MSFDA methods (DECISION, CAiDA and KD3A), which leverage state-of-the-art model selection methods (ANE, NMI, MDE, LEEP$^*$, and LogME$^*$) for source models selection before applying the MSFDA methods. The comparison of our method with these methods are shown in Table \ref{tab_home_2}, Table \ref{tab_31_2} and Table \ref{tab_dn_2}.
Our method  demonstrates effectiveness across all datasets consistently.
Compared with MSB-MSFDA methods, our approach achieves an improvement of 2.1\%, 2.5\% and 19.2\% on the Office-Home, Office-31, and DomainNet datasets, respectively.

\input{table/bench_s3}
\input{table/bench_s4}
\input{table/bench_s5}
\input{table/bench_s6}

\begin{figure*}
  \centering
 
  \includegraphics[width=0.95\linewidth]{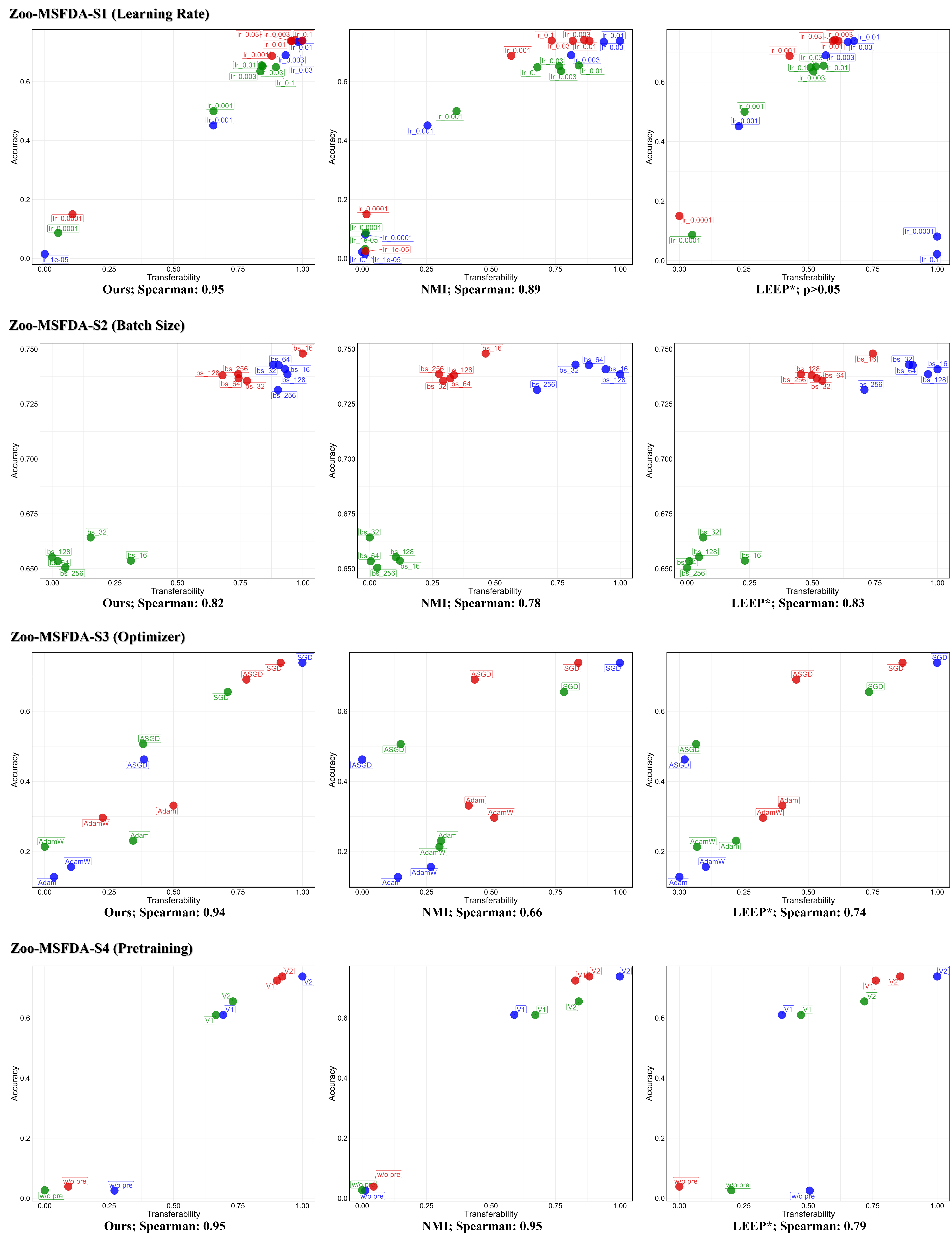}
\caption{The joint distribution (in the Zoo-MSFDA-S1/S2/S3/S4 setting) of 1) the performance of source models on target domain, and 2) the transferability of source models to the target domain estimated by using different methods. The Spearman's Rank Correlation Coefficient is used as the evaluation metric. The target domain is Office-Home's Re. Source models are from source domains including: Ar (blue), Cl (green), and Pr (red), with the same architecture (Resnet50). $p$-value $>$0.05 indicates the results do not have statistics significance.
}
\label{fig:joint_Zoo-MSFDA_s3456}
\end{figure*}

\subsection{Experiment\#3: Transferability Estimation Analysis}\label{sec:exp3}

\textbf{\textit{Comparison with Target-Only Transferability Estimation Methods.}}
Here we evaluate the effectiveness of our SUTE, the proposed transferability measurements. The compared methods include ANE, NMI, MDE, LEEP$^*$ and LogME$^*$. Results are shown in Table \ref{tab_home_tran}, Table \ref{tab_31_tran} and Table \ref{tab_dn_tran}. 
We can see that existing transferability measurements exhibit considerable instability in the Zoo-MSFDA setting where all domain shift, absence of source data, and architectural differences simultaneously occur. A case in point is LEEP$^*$. While it demonstrates an average correlation of 0.84 on the Office-Home dataset, no significant correlation is observed on the DomainNet dataset, as evidenced by $p$-value $>$ 0.05. This phenomenon explains the performance degradation when applying it to assist MSFDA methods on the DomainNet dataset (Table \ref{tab_dn_2}).
In contrast, our SUTE exhibits a strong correlation with the performance of source models across all datasets. It achieves a correlation of 0.94/0.97/0.89 on the Office-Home/Office-31/DomainNet dataset, surpassing the second-best method by 0.10/0.28/0.17.

We then compare our method with SND and MixVal, which  implicitly assume uniform model architectures. As shown in Table \ref{tab_home_tran_ora}, the two methods are not effective in the Zoo-MSFDA setting with architectural differences among models. Compared with these methods, our SUTE demonstrates greater effectiveness with an improvement of 0.14.

\noindent\textbf{\textit{Comparison with Oracle Transferability Estimation Methods.}}
We further consider some oracle methods that require target labels or source data. These methods include: standard LEEP and standard LogME that require target labels, and Maximum Mean Discrepancy (MMD) and A-Distance that requires source data. The experiments are conducted on Office-Home task. Results are reported in Table \ref{tab_home_tran_ora}. 
Surprisingly, we find that conventional UDA methods such as MMD and A-Distance do not perform well in transferability estimation tasks. As shown in Table \ref{tab_home_tran_ora}, MMD loses efficiency in tasks A, C, P$\rightarrow$R; A, P, R$\rightarrow$C; and C, P, R$\rightarrow$A. Similarly, A-Distance struggles in tasks A, C, R$\rightarrow$P and A, P, R$\rightarrow$C.

From Table \ref{tab_home_tran_ora}, it is evident that LEEP and LogME (both requiring target labels) demonstrate strong correlations (0.90 and 0.85, respectively) with the performance of source models on the target domain, indicating their effectiveness. 
Remarkably, \textit{without requiring target labels or source data}, our method demonstrates superior performance than these oracle methods by 0.04 (Ours: 0.94; LogME: 0.90).  The improvement achieved with fewer requirements further demonstrates the effectiveness of our approach.

\noindent\textbf{\textit{Joint Distribution Visualization.}} In Fig. \ref{fig:joint_Zoo-MSFDA}, we draw the joint distribution of 1) the performance of source models on target domain, and 2) the estimated transferability of source models by using our SUTE and existing methods (LEEP$^*$, NMI). The figure provides intuitive evidence demonstrating the effectiveness of our SUTE against both domain shifts and architecture discrepancies simultaneously.

\subsection{Experiment\#4: Ablation Analysis}\label{sec:exp4}
\noindent \textbf{Effectiveness of the Transferability Principle.}
We replace the SUTE, which effectively applies this principle, by two less effective measurements: NMI and LEEP$^*$. The two new approaches are denoted as Ours-NMI and Ours-LEEP$^*$, respectively.
Experiment results are reported in Tables \ref{tab_home_2}, Tables \ref{tab_31_2} and \ref{tab_dn_2}. Specifically, by replacing SUTE with NMI, the performance degraded by 6.1\%, 2.9\% and 17.8\% on the Office-Home, Office-31 and DomainNet datasets, respectively. By replacing SUTE with LEEP$^*$, the performance degraded by 2.0\%, 26.7\% and 36.4\% on the Office-Home, Office-31 and DomainNet datasets, respectively. These results demonstrate the importance of accurately selecting transferable source models in Zoo-MSFDA, thus highlighting the effectiveness of the transferability principle.

\noindent \textbf{Effectiveness of IC, SC and GD.}
In Table \ref{tab_home_tran}, Table \ref{tab_31_tran} and Table \ref{tab_dn_tran}, we evaluate the performance of SUTE without IC, SC and GD, respectively. The results demonstrate the effectiveness of all three characteristics. In addition, it becomes evident that SC and GD are of paramount importance in determining the performance of SUTE.

\noindent \textbf{Effectiveness of the Greedy Strategy.}
We introduce a greedy strategy that automatically collects the transferable set $S_{TR}$. 
To evaluate its effectiveness, we compare it with the Top$_n$ selection strategy, where we select the Top$_n$ most transferable models to establish $S_{TR}$, with $n$ set to 1/2/3/5/10, respectively. The results can be found in the Supplementary Table 14. The results demonstrate that the greedy strategy achieves the best performance on most task. Moreover, since this strategy does not involve hyperparameter, it is better suited for varying real-world scenarios.

\noindent \textbf{Effectiveness of the Diversity Principle.}
Here we evaluate the effectiveness of the diversity principle. To this end, we directly use the models in the selected transferable set (i.e., $S_{TR}$ in Fig. \ref{fig:method}) as the inlier models $S_I$. We term this produce as Our-w/oDiv. Its performance is reported in Table \ref{tab_home_2}, Table \ref{tab_31_2} and Table \ref{tab_dn_2}. It can be seen that Ours surpasses Our-w/oDiv by 1.4\%, 0.4\% and 1.4\% on the Office-Home, Office-31 and DomainNet datasets. The result demonstrates the importance of the diversity principle. Moreover, we observe that the diversity principle yields greater improvements when the transferability of source models is low, as seen on the Office-Home and DomainNet datasets. This suggests that the diversity principle may complement the transferability principle.

\noindent \textbf{Effectiveness of the Inlier Model Ensemble.}
We introduce a SUTE-based weights to ensemble inlier models rather than based on learnable weights as previous done \cite{dong2021confident,ahmed2021unsupervised}. 
We evaluate the effectiveness of this strategy in Table \ref{tab_home_2}, Table \ref{tab_31_2} and Table \ref{tab_dn_2}, where Ours-lw denotes the use of learnable weights. We can see that with SUTE-based weights, Ours outperforms Ours-lw in all Office-Home, Office-31 and DomainNet datasets.

\noindent \textbf{Effectiveness of the Outlier Model Recycle.}
In Ours-w/o$\mathscr{L}_{omr}$ of Table \ref{tab_home_2} and Table \ref{tab_dn_2}, we removed the term $\mathscr{L}_{omr}$ from the overall objective. This adjustment led to a slight performance decrease of 0.3\%, 0.8\% and 0.7\% on the Office-Home, Office-31 and DomainNet datasets, respectively. This outcome indicates that the benefit of recycling knowledge from outlier models.
\noindent \textbf{Effectiveness of the Pseudo-Label Learning.}
In Ours-w/o$\mathscr{L}_{pse}$ of Table \ref{tab_home_2}, Table \ref{tab_31_2} and Table \ref{tab_dn_2}, we removed the term $\mathscr{L}_{pse}$ from the overall objective. This decreases the performance by 1.9\%, 2.6\% and 1.1\% on the Office-Home, Office-31 and DomainNet datasets, respectively.

\noindent \textbf{Effectiveness of the Separate Information Maximization.}
Here we compare the proposed separate information maximization $\mathscr{L}_{sim}$ and conventional information maximization in MSFDA (i.e., $\mathscr{L}_{cim}$). Ours-$\mathscr{L}_{cim}$ denotes that replacing $\mathscr{L}_{sim}$ by $\mathscr{L}_{cim}$ in the overall objective. Results are reported in Table \ref{tab_home_2}, Table \ref{tab_31_2} and \ref{tab_dn_2}. We can see that Ours outperforms Ours-$\mathscr{L}_{cim}$ by 0.5\%, 0.3\% and 0.5\% on the Office-Home, Office-31 and DomainNet datasets, which demonstrates the superior of $\mathscr{L}_{sim}$.

\subsection{Experiment\#5: Analysis on Sub-Settings}\label{sec:exp5}

\noindent \textbf{Performance in Zoo-MSFDA-S1.} This setting leverages multiple models with different learning rates. The performance is reported in Table \ref{tab_home_s3}. Our method achieves the highest performance, surpassing existing method by 2.0\%.
The transferability estimation of different methods in this settings is presented in Table \ref{tran_other_settings}.
ANE, NMI and MDE works well in this setting, while LogME$^*$ and LEEP$^*$ lost efficiency. Compared with these approaches, our method demonstrates a stronger correlation with the source models' performance in the target domain (0.91 \textit{vs.} 0.89).

\noindent \textbf{Performance in Zoo-MSFDA-S2.} This setting leverages multiple models with different batch sizes. The performance is reported in Table \ref{tab_home_s4}. Our method achieves 72.6\% classification accuracy, which is competitive with existing methods.
The transferability estimation of different methods in this settings is presented in Table \ref{tran_other_settings}.
ANE, NMI, LogME$^*$ and LEEP$^*$ show strong correlation in this setting (0.84, 0.84, 0.81, and 0.88, respectively). Compared with them, our method demonstrates competitive correlation (0.85).

\noindent \textbf{Performance in Zoo-MSFDA-S3.} This setting leverages multiple models with different optimizers. The performance is reported in Table \ref{tab_home_s5}. 
Our method achieves a performance of 64.7\%, surpassing CAiDA-based methods and DECISION-based methods. However, it falls short of the performance attained by KD3A+NMI (66.2\%). We attribute this disparity to the diminished efficiency of the greedy strategy in task A, P, R$\rightarrow$C. As shown in Table \ref{tab_home_s5}, by adopting the Top$_2$ selection strategy (directly choosing the Top$n$ most transferable models to form $S{TR}$), we achieve the highest performance of 68.1\% in the Zoo-MSFDA-S3.

The transferability estimation of different methods in this settings is presented in Table \ref{tran_other_settings}.
Our method achieves 0.79 correlation, significantly surpassing existing methods by 0.21.

\noindent \textbf{Performance in Zoo-MSFDA-S4.} This setting leverages multiple models initialized by different pre-trained weights. The performance is reported in Table \ref{tab_home_s5}.  Our method achieves superior performance, surpassing existing methods by 2.6\%.
The transferability estimation of different methods in this setting is presented in Table \ref{tran_other_settings}. It is observed that ANE and NMI exhibit strong correlations in this setting (0.96 and 0.97, respectively), while our method demonstrates a competitive correlation of 0.96.

\noindent \textbf{Joint Distribution Visualization.}
In Fig. \ref{fig:joint_Zoo-MSFDA_s3456}, we present the joint distribution of source models' true performance and their transferability estimated by LEEP$^*$, NMI, and our SUTE in the Zoo-MSFDA-S1/S2/S3/S4. 
It reveals that our method consistently performs effectively across all tasks. These results emphasize the robustness of our approach against various training configurations, encompassing learning rate, batch size, optimizer, and pre-trained weights.

\subsection{Hyperparameter Analysis}
We conducted hyperparameter analysis to demonstrate the stability of hyperparameters. The details and results are shown in Supplementary 4.2.

%% file: table/tab_home_1.tex
\begin{table}\scriptsize
\caption{Performance (\%) comparison between our method and existing MSFDA methods  on the Office-Home dataset. $\rightarrow$R denotes the tasks A, C, P$\rightarrow$R. ``Avg." denotes the average accuracy. ``Ours-tf": using the ensemble of source models selected by our selection procedure without adaptation.}
\centering
\begin{tabular}{|l|l|c|c|c|c|c|}
\hline
Setting & Method & $\rightarrow$R & $\rightarrow$P &$\rightarrow$C & $\rightarrow$A & Avg. \\
\hline
 \multirow{4}{*}{MSFDA}& Source & 76.3 & 78.8 & 50.1 & 50.9 & 64.0 \\
   
 & DECISION & 83.6 & 84.4 & 59.4 & 74.5 & 75.5 \\
 & CAiDA & 84.2 & 84.7 & 60.5 & 75.2 & 76.2 \\
  & KD3A & 83.8 & 84.0 & 58.7 & 74.1 & 75.2 \\ 
\hline
 \multirow{6}{*}{Zoo-MSFDA}& Source & 80.4 & 74.8 & 52.1 & 64.2 & 69.5 \\
 
 & DECISION & 90.1 & 88.9 & 65.3 & 82.4 & 81.6 \\
 & CAiDA & 89.2 & 86.7 & 65.6 & 82.7 & 81.1 \\
 & KD3A  &90.6 &89.4 & 65.8&82.5 & 82.1 \\
& Ours-tf & 93.6 & 93.8 & 83.2 & 92.3 & 90.7 \\

 & Ours & 95.2 & 95.3 & 83.7 & 92.8 & 91.8 \\
\hline

\end{tabular}
\label{tab_home_1}
\end{table}

%% file: table/tab_31_1.tex
\begin{table}\scriptsize
\caption{Performance (\%) comparison between our method and existing MSFDA methods on the Office-31 dataset. $\rightarrow$W denotes the tasks A, D$\rightarrow$W. ``Avg." denotes the average accuracy. ``Ours-tf": using the ensemble of source models selected by our selection procedure without adaptation.}
\centering
\begin{tabular}{|l|l|c|c|c|c|}
\hline
Setting & Method & $\rightarrow$W & $\rightarrow$D &$\rightarrow$A & Avg. \\
\hline
 \multirow{4}{*}{MSFDA}& Source & 76.3 & 78.8 & 50.1 & 50.9  \\
   
 &DECISION$^\star$  &98.4 &99.6& 75.4& 91.1\\
 &CAiDA$^\star$ &98.9 &99.8 &75.8 &91.6\\
 &KD3A$^\star$ &98.3 &99.0 &75.1 &91.6\\
\hline
 \multirow{6}{*}{Zoo-MSFDA}& Source & 72.6 & 83.1 & 50.0  &  68.6 \\
 & DECISION &99.1 & 99.2 &  84.3&94.2\\
 & CAiDA &97.8 & 99.0 & 83.5 &93.4 \\
& KD3A &97.7 & 99.4 &80.7 & 92.6\\
& Ours-tf &98.7 & 99.8 & 81.7 &93.4 \\

 & Ours &99.1  & 100.0 & 85.6 &94.9 \\
\hline

\end{tabular}
\label{tab_31_1}
\end{table}

%% file: table/tab_dn_1.tex
\begin{table}\scriptsize
\caption{Performance (\%) comparison between our method and existing MSFDA methods on the DomainNet dataset. $\rightarrow$C denotes the transfer task I, P, Q, R, S$\rightarrow$C. ``Avg." denotes the average accuracy. ``Ours-tf": using the ensemble of source models selected by our selection procedure without adaptation.}
\centering
\resizebox{\linewidth}{!}{
\begin{tabular}{|l|l|c|c|c|c|c|c|c|}
\hline
Setting & Method &$\rightarrow$C & $\rightarrow$I &$\rightarrow$P &$\rightarrow$Q & $\rightarrow$R &$\rightarrow$S & Avg. \\
\hline
  \multirow{4}{*}{MSFDA}& Source & 49.3 & 14.2 & 39.4 & 12.6 & 53.0 & 35.1 & 33.9 \\
 & DECISION & 61.5 & 21.6 & 54.6 & 18.9 & 67.5 & 51.0 & 45.9 \\
 & CAiDA & 63.6 & 20.7 & 54.3 & 19.3 & 71.2 & 51.6 & 46.8 \\
 & KD3A & 69.7 & 21.2 & 58.8& 15.1 & 70.4 & 57.9 & 48.8 \\
\hline
 \multirow{5}{*}{Zoo-MSFDA} & Source & 40.8 & 17.1 & 36.2 & 8.2 & 47.8 & 33.3 & 30.5 \\
 & DECISION &43.0&17.8&43.4&13.2&59.6&55.4&38.7 \\
 & KD3A &41.6&17.8&43.1&15.1&58.9&56.5&38.8 \\
 &Ours-tf&79.9&38.6&70.1&14.1&82.7&69.7&59.1\\
 & Ours & 80.5 & 48.5 & 72.9 & 17.9 & 83.8 & 69.9 & 62.3 \\
\hline
\end{tabular}
}
\label{tab_dn_1}
\end{table}

%% file: table/tab_home_2.tex
\begin{table}\scriptsize
\caption{Performance (\%) comparison between our method and MSB-MSFDA methods on the Office-Home dataset in the Zoo-MSFDA setting. ``D+/C+/K+'' represents the adaptation method DECISION/CAiDA/KD3A. ``ANE/NMI/MDE/LEEP$^*$/LogME$^*$'' represents the selection method. $\rightarrow$R denotes the tasks A, C, P$\rightarrow$R. ``Avg." denotes the average accuracy. ``Ours-NMI" and ``Ours-LEEP$*$" denote replacing the SUTE loss with the NMI and LEEP$*$ losses, respectively. ``Ours-w/oDiv" excludes the Diversity set for model selection, while ``Ours-lw" utilizes learnable weights instead of SUTE-based weights for ensemble. ``Ours-w/o$\mathscr{L}_{pse}$" and ``Ours-w/o$\mathscr{L}_{omr}$" evaluate the performance without  $\mathscr{L}_{pse}$ and $\mathscr{L}_{omr}$, respectively.
``Ours-$\mathscr{L}_{cim}$" replaces the $\mathscr{L}_{sim}$ with the $\mathscr{L}_{cim}$. ``-" indicates significantly low performance (accuracy $<$5.0\%).}
\centering
\begin{tabular}{|l|c|c|c|c|c|}
\hline
 Method & $\rightarrow$R & $\rightarrow$P &$\rightarrow$C & $\rightarrow$A & Avg. \\
\hline

 D+ANE & 89.9 & 92.6 & 66.8 & 84.8 & 83.0 \\
 D+NMI & 89.9 & 92.0 & 69.9 & 85.6 & 84.3 \\
 D+LEEP$^*$ & 91.9 & 92.7 & 81.4 & 90.2 & 89.1 \\
 D+LogME$^*$ & 90.2 & 91.0 & 66.3 & 83.4 & 82.7 \\
 D+MDE & 86.5 & 86.5 & 64.0 & 79.2 & 79.1 \\
\hline

 C+ANE & 89.7 & 92.5 & 65.8 & 83.7 & 82.9 \\
 C+NMI & 88.7 & 92.3 & 69.5 & 84.3 & 83.7 \\
 C+LEEP$^*$ & 92.1 & 92.3 & 81.3 & 90.4 & 89.0 \\
 C+LogME$^*$ & 89.9 & 89.1 & 64.3 & 82.9 & 82.0 \\
 C+MDE & 85.8 & 86.6 & 63.7 & 79.2 & 78.8 \\
 
\hline

 K+ANE  & 90.2& 91.3&66.7 & 82.4&83.1  \\
 K+NMI  & 90.3&91.0 &70.1 &83.9 &83.8  \\
 K+LEEP$^*$ &92.2 &93.9 &82.0&90.7 & 89.7 \\
  K+LogME$^*$  &90.3& 88.9&67.5 & 82.5&82.3  \\
 K+MDE & 85.9 & 83.1 & 61.9 & 78.0 & 77.2 \\

\hline
 Ours-NMI & 90.8 & 92.4 & 72.5 & 87.4 & 85.7 \\
 Ours-LEEP$^*$ & 92.2 & 93.9 & 82.4 & 90.9 & 89.8 \\
 Ours-w/oDiv & 94.0 & 93.8 & 82.9 & 91.2 & 90.4 \\
Ours-lw & 94.6 & 94.1 & 83.2 & 92.3 & 91.1 \\
 Ours-w/o$\mathscr{L}_{pse}$ & 93.7 & 93.1 & 82.0 & 90.9 & 89.9 \\
 Ours-w/o$\mathscr{L}_{omr}$ & 95.1 & 94.9 & 83.4 & 92.6 & 91.5 \\
 Ours-$\mathscr{L}_{cim}$ & 94.8 & 94.8 & 83.2 & 92.5 & 91.3 \\
 Ours & 95.2 & 95.3 & 83.7 & 92.8 & 91.8 \\
\hline

\end{tabular}
\label{tab_home_2}
\end{table}

%% file: table/tab_31_2.tex
\begin{table}\scriptsize
\caption{Performance (\%) comparison between our method and MSB-MSFDA methods on the Office-31 dataset in the Zoo-MSFDA setting. ``D+/C+/K+'' represents the adaptation method DECISION/CAiDA/KD3A. ``ANE/NMI/MDE/LEEP$^*$/LogME$^*$'' represents the selection method. $\rightarrow$W denotes the tasks A, D$\rightarrow$W. ``Avg." denotes the average accuracy. ``Ours-NMI" and ``Ours-LEEP$*$" denote replacing the SUTE loss with the NMI and LEEP$*$ losses, respectively. ``Ours-w/oDiv" excludes the Diversity set for model selection, while ``Ours-lw" utilizes learnable weights instead of SUTE-based weights for ensemble. ``Ours-w/o$\mathscr{L}_{pse}$" and ``Ours-w/o$\mathscr{L}_{omr}$" evaluate the performance without  $\mathscr{L}_{pse}$ and $\mathscr{L}_{omr}$, respectively.
``Ours-$\mathscr{L}_{cim}$" replaces the $\mathscr{L}_{sim}$ with the $\mathscr{L}_{cim}$. ``-" indicates significantly low performance (accuracy $<$5.0\%).}
\centering
\begin{tabular}{|l|c|c|c|c|c|}
\hline
 Method & $\rightarrow$W & $\rightarrow$D &$\rightarrow$A & Avg. \\
\hline


 D+ANE&97.9 &98.4&78.0&91.3 \\
 D+NMI &97.8 &98.4&77.9&91.4 \\
 D+LEEP$^*$ &64.5 &87.1&17.9&56.5 \\
 D+LogME$^*$&96.7 &99.0&60.2&85.3 \\
 D+MDE &97.7 &98.8&78.6&91.7 \\
\hline

 C+ANE&98.0 &99.4&80.0&92.4 \\
 C+NMI &97.7 &99.2&78.8&91.9 \\
 C+LEEP$^*$&64.1 &87.2&17.9&56.4 \\
 C+LogME$^*$ & 96.9&98.8&60.6&85.4 \\
 C+MDE &97.0 &98.0&82.3&92.4 \\
 
\hline

 K+ANE  &96.3 &97.6&78.6&90.8 \\
 K+NMI &96.3&97.5&78.6&90.8 \\
 K+LogME$^*$ &93.3 &98.7&67.1&86.3 \\
 K+LEEP$^*$& 72.6&99.2&77.0&82.9 \\
 k+MDE&96.2 &97.1&77.6&90.3 \\

\hline
 Ours-NMI &97.8 &99.1&79.2&92.0 \\
 Ours-LEEP$^*$&80.1 &96.7&27.9& 68.2\\
 Ours-w/oDiv &98.7&100.0&84.8&94.5 \\
 Ours-lw &98.4&100.0&85.2&94.5 \\
 Ours-w/o$\mathscr{L}_{pse}$ &98.4 &99.2&82.5&92.3 \\
 Ours-w/o$\mathscr{L}_{omr}$&98.7 &100.0&83.7&94.1 \\
 Ours-$\mathscr{L}_{cim}$& 99.1&100.0&84.9&94.6 \\
 Ours &99.1  & 100.0 & 85.6 &94.9 \\
\hline

\end{tabular}
\label{tab_31_2}
\end{table}

%% file: table/tab_dn_2.tex
\begin{table}\scriptsize
\caption{Performance (\%) comparison between our method and MSB-MSFDA methods on the DomainNet dataset in the Zoo-MSFDA setting.  ``D+/K+'' represents the adaptation method DECISION/KD3A. ``ANE/NMI/MDE/LEEP$^*$/LogME$^*$'' represents the selection method.
$\rightarrow$C denotes the transfer task I, P, Q, R, S$\rightarrow$C. ``Avg." denotes the average accuracy. ``Ours-NMI" and ``Ours-LEEP$*$" denote replacing the SUTE loss with the NMI and LEEP$*$ losses, respectively. ``Ours-w/oDiv" excludes the Diversity set for model selection, while ``Ours-lw" utilizes learnable weights instead of SUTE-based weights for ensemble. ``Ours-w/o$\mathscr{L}_{pse}$" and ``Ours-w/o$\mathscr{L}_{omr}$" evaluate the performance without  $\mathscr{L}_{pse}$ and $\mathscr{L}_{omr}$, respectively.
``Ours-$\mathscr{L}_{cim}$" replaces the $\mathscr{L}_{sim}$ with the $\mathscr{L}_{cim}$. ``-" indicates significantly low performance (accuracy $<$5.0\%).}
\centering
\begin{tabular}{|l|c|c|c|c|c|c|c|}
\hline
 Method &$\rightarrow$C & $\rightarrow$I &$\rightarrow$P &$\rightarrow$Q & $\rightarrow$R &$\rightarrow$S & Avg. \\
\hline

 D+ANE & 40.2 & 11.2 & 41.6 & 5.1 & 78.6 & 57.3 & 39.0 \\
 D+NMI & 41.4 & 15.2 & 43.6 & 18.4 & 78.8 & 60.1 & 43.0 \\
 D+LogME$^*$ &60.6& 17.0 &30.2&-& 60.4 &10.5&N/A \\
 D+LEEP$^*$ & 40.3 & 7.7 & 13.8 & 9.7 & 18.5 & 36.8 & 21.1 \\
 D+MDE & - & - & - & 12.8 & - & - & N/A \\
 \hline
 
 K+ANE &37.2&17.1&37.8&-& 77.4 &57.5& N/A\\
 K+NMI &40.5&16.3&40.5&19.0& 79.9 &62.8&43.1 \\
 
 K+LEEP$^*$ &23.0&-&13.4&10.5& 17.8 &30.1&N/A \\
 K+LogME$^*$ &59.1&-&29.1&-& 67.0 &13.5&N/A \\
 K+MDE &28.2& 8.8 &26.1&12.1& 40.4 &51.4&27.8 \\
\hline
 Ours-NMI & 43.5 & 18.2 & 46.5 & 17.9 & 79.1 & 61.8 & 44.5 \\
 Ours-LEEP$^*$ & 41.3 & 16.7 & 18.1 & 16.7 & 24.9 & 38.2 & 25.9 \\
 Ours-w/oDiv & 79.0 & 47.7 & 71.3 & 16.8 & 83.2 & 68.0 & 60.9 \\
Ours-lw & 79.3 & 47.1 & 72.1 & 17.6 & 83.7 & 68.8 & 61.4 \\
 Ours-w/o$\mathscr{L}_{pse}$ & 79.2 & 47.4 & 71.9 & 17.3 & 83.0 & 68.5 & 61.2 \\
 Ours-w/o$\mathscr{L}_{omr}$ & 79.4 & 47.7 & 72.3 & 17.1 & 83.6 & 69.2 & 61.6 \\
 Ours-$\mathscr{L}_{cim}$ & 80.1 & 48.0 & 72.6 & 17.7 & 83.4 & 69.5 & 61.8 \\
 Ours & 80.5 & 48.5 & 72.9 & 17.9 & 83.8 & 69.9 & 62.3 \\
\hline
\end{tabular}
\label{tab_dn_2}
\end{table}

%% file: table/tab_home_tran.tex
\begin{table}\scriptsize
\caption{Spearman's Rank Correlation Coefficient between the performance and the measured transferability of source models on the target domain in the Zoo-MSFDA setting. Experiments are conducted on the Office-Home dataset. $\rightarrow$R denotes the tasks A, C, P$\rightarrow$R. ``-" denotes that the results do not have statistics significance (i.e., $p$-value $>$0.05).}
\centering
\begin{tabular}{|l|c|c|c|c|c|c|c|}
\hline
 Method & $\rightarrow$R & $\rightarrow$P &$\rightarrow$C & $\rightarrow$A & Avg. \\
\hline
ANE&0.52&0.74&0.57&0.73&0.64\\
NMI&0.52&0.76&0.64&0.74&0.67\\
LogME$^*$&0.79&0.76&0.81&0.82&0.80\\
LEEP$^*$&0.79&0.89&0.78&0.90&0.84\\
MDE&-&-&0.42&0.39&N/A\\
\hline
SUTE-w/o IC &0.92&0.95&0.88&0.95&0.93\\
SUTE-w/o SC &0.58&0.80&0.70&0.81&0.72\\
SUTE-w/o GD &0.93&0.94&0.86&0.95&0.92\\
SUTE (Ours)&0.94&0.97&0.88&0.97&0.94\\
\hline
\end{tabular}
\label{tab_home_tran}
\end{table}

%% file: table/tab_31_tran.tex
\begin{table}\scriptsize
\caption{Spearman's Rank Correlation Coefficient between the performance and the measured transferability of source models on the target domain in the Zoo-MSFDA setting. Experiments are conducted on the Office-31 dataset.  $\rightarrow$W denotes the tasks A, D$\rightarrow$W. ``-" denotes that the results do not have statistics significance (i.e., $p$-value $>$0.05).}
\centering
\begin{tabular}{|l|c|c|c|c|c|c|}
\hline
 Method & $\rightarrow$W & $\rightarrow$D &$\rightarrow$A & Avg. \\
\hline
ANE&0.85&0.50&0.67&0.67\\
NMI&0.86&0.51&0.70&0.69\\
LogME$^*$&-&-&-&N/A\\
LEEP$^*$&-&-&-&N/A\\
MDE&0.77&0.38&0.41&0.52\\
\hline
SUTE-w/o IC &0.96&0.93&0.97&0.95\\
SUTE-w/o SC &0.67&0.37&0.78&0.60\\
SUTE-w/o GD &0.96&0.96&0.98&0.97\\
SUTE (Ours)&0.96&0.97&0.98&0.97\\
\hline
\end{tabular}
\label{tab_31_tran}
\end{table}

%% file: table/tab_dn_tran.tex
\begin{table}\scriptsize
\caption{Spearman's Rank Correlation Coefficient between the performance and the measured transferability of source models on the target domain in the Zoo-MSFDA setting. Experiments are conducted on the DomainNet dataset.  $\rightarrow$R denotes the tasks A, C, P$\rightarrow$R. ``-" denotes that the results do not have statistics significance (i.e., $p$-value $>$0.05).}
\centering
\begin{tabular}{|l|c|c|c|c|c|c|c|}
\hline
 Method &$\rightarrow$C & $\rightarrow$I &$\rightarrow$P &$\rightarrow$Q & $\rightarrow$R &$\rightarrow$S & Avg. \\
\hline
ANE&0.61&-&0.57&-&0.52&0.66&N/A\\
NMI&0.73&0.62&0.73&0.77&0.69&0.78&0.72\\
LogME$^*$&-&0.48&0.58&-&0.33&-&N/A\\
LEEP$^*$&-&-&-&-&-&-&N/A\\
MDE&-&-&-&-&0.22&-&N/A\\
\hline
SUTE-w/o IC &0.91&0.77&0.89&0.86&0.90&0.92&0.87\\
SUTE-w/o SC &0.73&-&0.74&0.79&0.70&0.78&N/A\\
SUTE-w/o GD &-&-&-&0.55&-&-&N/A\\
SUTE(Ours)&0.93&0.77&0.89&0.86&0.91&0.95&0.89\\
\hline
\end{tabular}
\label{tab_dn_tran}
\end{table}

%% file: table/tab_home_tran_ora.tex
\begin{table}\scriptsize
\caption{Spearman's Rank Correlation Coefficient between the performance and the measured transferability of source models on the target domain in the Zoo-MSFDA setting. Experiments are conducted on the Office-Home dataset.  $\rightarrow$R denotes the tasks A, C, P$\rightarrow$R. ``TL"/``SD" indicates the requirement of target label/source data. ``-" denotes that the results do not have statistics significance (i.e., $p$-value $>$0.05).}
\centering
\begin{tabular}{|l|c|c|c|c|c|c|c|}
\hline
 Method &TL&SD& $\rightarrow$R & $\rightarrow$P &$\rightarrow$C & $\rightarrow$A & Avg. \\
\hline

LogME&\checkmark&$\times$&0.85&0.92&0.87&0.94&0.90\\
LEEP&\checkmark&$\times$&0.84&0.80&0.93&0.84&0.85\\
MMD&$\times$&\checkmark&-&0.26&-&-&N/A\\
A-Distance&$\times$&\checkmark&-&-&-&-&N/A\\
\hline
SND&$\times$&$\times$&0.83&0.75&0.85&0.76&0.80\\
MixVal&$\times$&$\times$&0.25&-&-&-&N/A\\
\hline
SUTE(Ours)&$\times$&$\times$&0.94&0.97&0.88&0.97&0.94\\
\hline
\end{tabular}
\label{tab_home_tran_ora}
\end{table}

%% file: table/tab_other_tran.tex
\begin{table}\scriptsize
\caption{Spearman's Rank Correlation Coefficient between the performance and the measured transferability of source models on the target domain in different sub-settings. Experiments are conducted on the Office-Home dataset. $\rightarrow$R denotes the tasks A, C, P$\rightarrow$R. ``-" denotes that the results do not have statistics significance (i.e., $p$-value $>$0.05).}
\centering
\begin{tabular}{|l|l|c|c|c|c|c|}
\hline

Setting&Method&$\rightarrow$R &$\rightarrow$P &$\rightarrow$C&$\rightarrow$A&Avg.\\

\hline


\hline
\multirow{6}{*}{Zoo-MSFDA-S1}
&ANE&0.84&0.87&0.84&0.86&0.85\\
&NMI&0.89&0.91&0.87&0.88&0.89\\
&LogME$^*$&-&-&-&-&N/A\\
&LEEP$^*$&-&-&0.20&-&N/A\\
&MDE&0.82&0.81&0.66&0.80&0.77\\
&SUTE(Ours)&0.95&0.84&0.93&0.92&0.91\\
\hline
\hline
\multirow{6}{*}{Zoo-MSFDA-S2}
&ANE&0.78&0.94&0.70&0.94&0.84\\
&NMI&0.78&0.95&0.69&0.95&0.84\\
&LogME$^*$&0.87&0.94&0.92&0.51&0.81\\
&LEEP$^*$&0.83&0.96&0.81&0.91&0.88\\
&MDE&0.56&0.66&0.56&0.90&0.67\\
&SUTE(Ours)&0.82&0.93&0.86&0.77&0.85\\
\hline
\hline
\multirow{6}{*}{Zoo-MSFDA-S3}
&ANE&0.58&0.84&0.28&0.41&0.53\\
&NMI&0.66&0.84&0.38&0.44&0.58\\
&LogME$^*$&-&-&-&-&N/A\\
&LEEP$^*$&0.74&0.73&0.44&-&N/A\\
&MDE&-&0.24&-&-&N/A\\
&SUTE(Ours)&0.94&0.92&0.52&0.76&0.79\\
\hline
\hline\multirow{6}{*}{Zoo-MSFDA-S4}
&ANE&0.95&0.95&0.95&0.98&0.96\\
&NMI&0.95&0.98&0.95&0.98&0.97\\
&LogME$^*$&-&-&-&-&N/A\\
&LEEP$^*$&0.79&0.58&0.58&0.58&0.63\\
&MDE&0.86&0.92&-&0.93&N/A\\
&SUTE(Ours)&0.95&0.98&0.90&1.00&0.96\\
\hline

\end{tabular}\label{tran_other_settings}
\end{table}

%% file: table/bench_s3.tex
\begin{table}\scriptsize
\caption{Performance (\%) on the Office-Home dataset in the Zoo-MSFDA-S1 setting. $\rightarrow$R denotes the tasks A, C, P$\rightarrow$R. ``-" indicates significantly low performance (accuracy $<$5.0\%).}
\centering
\begin{tabular}{|l|c|c|c|c|c|}

\hline
Method & $\rightarrow$R & $\rightarrow$P &$\rightarrow$C & $\rightarrow$A & Avg. \\
\hline
DECISION&75.4&76.1&40.7&66.9&64.8\\
+ANE&79.3&78.8&-&64.7&N/A\\
+NMI&79.9&78.2&49.3&66.1&68.4\\
+LogME$^*$&9.3&-&-&64.4&N/A\\
+LEEP$^*$&-&5.8&-&57.5&N/A\\
+MDE & 80.5& 79.6 & 49.4 & 66.3 & 68.9 \\
\hline
CAiDA&77.4&76.2&-&-&N/A\\
+ANE&78.5&77.5&-&-&N/A\\
+NMI&79.1&78.9&-&64.6&N/A\\
+LogME$^*$&-&-&-&-&N/A\\
+LEEP$^*$&-&11.2&-&57.5&N/A\\
+MDE & 80.4 & 79.3 & 43.7 & 66.0 & 67.4 \\
\hline
KD3A&79.3&77.4&48.4&66.5&67.9\\
+ANE&79.4&77.5&48.4&66.6&68.0\\
+NMI&79.9&77.5&48.4&67.0&68.2\\
+LogME$^*$&7.9&-&-&66.9&N/A\\
+LEEP$^*$&-&9.3&-&54.1&N/A\\
+MDE & 80.0 & 78.1 & 48.2 & 67.7 & 68.5 \\
\hline
Ours&82.5&80.3&51.6&69.3&70.9\\
\hline
\end{tabular}\label{tab_home_s3}
\end{table}

%% file: table/bench_s4.tex
\begin{table}\scriptsize
\caption{Performance (\%) on the Office-Home dataset in the Zoo-MSFDA-S2 setting. $\rightarrow$R denotes the tasks A, C, P$\rightarrow$R. ``-" indicates significantly low performance (accuracy $<$5.0\%).}
\centering
\begin{tabular}{|l|c|c|c|c|c|}

\hline
Method & $\rightarrow$R & $\rightarrow$P &$\rightarrow$C & $\rightarrow$A & Avg. \\
\hline
DECISION&81.4&80.6&51.7&68.8&70.6\\
+ANE/NMI/LogME$^*$&81.8&80.9&53.5&68.9&71.3\\
+LEEP$^*$&82.2&81.5&55.7&70.7&72.5\\
+MDE & 81.7 & 80.7 & 54.0 & 69.4 & 71.5 \\
\hline
CAiDA&80.6&80.7&52.2&67.0&70.1\\
+ANE/NMI/LogME$^*$&81.0&80.3&52.7&66.9&70.2\\
+LEEP$^*$&82.0&81.9&55.7&70.4&72.5\\
+MDE & 81.1 & 80.3 & 53.7 & 67.7 & 70.7 \\
\hline
KD3A&80.8&77.9&50.7&68.4&69.5\\
+ANE/NMI/LogME$^*$&80.8&77.9&50.7&68.4&69.5\\
+LEEP$^*$&81.2&79.1&52.1&68.6&70.3\\
+MDE & 79.6 & 77.9 & 50.7 & 68.9 & 69.3 \\
\hline
Ours&82.8&81.4&55.5&71.3&72.6\\
\hline
\end{tabular}\label{tab_home_s4}
\end{table}

%% file: table/bench_s5.tex
\begin{table}\scriptsize
\caption{Performance (\%) on the Office-Home dataset in the Zoo-MSFDA-S3 setting. $\rightarrow$R denotes the tasks A, C, P$\rightarrow$R. ``Ours(Top$_2$)'' denotes using the Top$_2$ most transferable models selected by our SUTE rather than using the greedy strategy. ``-" indicates significantly low performance (accuracy $<$5.0\%).}
\centering
\begin{tabular}{|l|c|c|c|c|c|}

\hline
Method & $\rightarrow$R & $\rightarrow$P &$\rightarrow$C & $\rightarrow$A & Avg. \\
\hline
DECISION&67.5&72.9&32.6&54.8&57.0\\
+ANE&67.9&73.6&33.2&56.5&57.8\\
+NMI&68.7&72.6&37.3&56.8&58.9\\
+LogME$^*$&46.6&-&24.2&10.7&N/A\\
+LEEP$^*$&79.3&-&50.6&10.7&N/A\\
+MDE & 64.2 & 69.6 & 31.8 & 51.3 & 54.2 \\
\hline
CAiDA&60.6&68.5&29.1&52.2&52.6\\
+ANE&65.9&72.3&29.0&-&N/A\\
+NMI&66.2&73.0&36.1&-&N/A\\
+LogME$^*$&-&-&-&10.4&N/A\\
+LEEP$^*$&79.4&-&50.9&10.6&N/A\\
+MDE & 62.8 & 68.8 & 32.4 & 50.6 & 53.7 \\
\hline
KD3A&76.3&74.7&45.8&63.2&65.0\\
+ANE&76.8&75.4&45.8&63.2&65.3\\
+NMI&77.2&75.8&46.5&65.1&66.2\\
+LogME$^*$&46.2&37.5&28.1&12.0&31.0\\
+LEEP$^*$&79.8&37.5&48.4&12.0&44.4\\
+MDE & 76.0 & 74.7 & 45.1 & 62.3 & 64.5 \\
\hline
Ours&80.8&77.4&34.1&66.6&64.7\\
Ours(Top$_2$)&80.3&77.0&49.8&65.2&68.1\\

\hline
\end{tabular}\label{tab_home_s5}
\end{table}

%% file: table/bench_s6.tex
\begin{table}\scriptsize
\caption{Performance (\%) on the Office-Home dataset in the Zoo-MSFDA-S4 setting. $\rightarrow$R denotes the tasks A, C, P$\rightarrow$R. ``-" indicates significantly low performance (accuracy $<$5.0\%).}
\centering
\begin{tabular}{|l|c|c|c|c|c|}

\hline
Method & $\rightarrow$R & $\rightarrow$P &$\rightarrow$C & $\rightarrow$A & Avg. \\
\hline
DECISION&76.1&73.1&35.6&61.4&61.6\\
+ANE&76.8&77.0&35.1&60.9&62.5\\
+NMI&77.0&77.7&46.0&62.1&65.7\\
+LogME$^*$&-&-&-&-&N/A\\
+LEEP$^*$&79.4&-&48.1&-&N/A\\
+MDE & 75.8 & 74.3 & 31.0 & 61.1 & 60.6 \\
\hline
CAiDA&76.4&75.9&37.3&59.6&62.3\\
+ANE&77.1&76.5&36.1&59.4&62.3\\
+NMI&77.1&78.6&45.9&62.1&65.9\\
+LogME$^*$&-&-&-&-&N/A\\
+LEEP$^*$&79.7&-&48.8&-&N/A\\
+MDE & 76.0 & 74.1 & 32.4 & 61.0 & 60.9 \\
\hline
KD3A&79.3&76.8&48.3&65.8&67.6\\
+ANE/NMI/MDE&79.3&77.0&48.3&65.8&67.6\\
+LogME$^*$&-&-&-&-&N/A\\
+LEEP$^*$&79.8&-&49.3&5.4&N/A\\
\hline
Ours&82.1&79.3&50.4&68.8&70.2\\
\hline
\end{tabular}\label{tab_home_s6}
\end{table}

%% file: conclusions.tex
\section{Conclusions}
In this paper, we introduce a new setting termed Zoo-MSFDA, which allows each source domain to offer a zoo trained source models with different architectures, as well as permits the target user to leverage any model from these model zoos without quantitative restrictions.
We provide theoretical analysis of the model selection problem in Zoo-MSFDA, and propose two principles (\textit{transferability principle} and \textit{diversity principle}) for appropriately selecting source models from models zoos. Further, we further propose a novel method named \textbf{S}ource-Free \textbf{U}nsupervised \textbf{T}ransferability \textbf{E}stimation (SUTE), which enables the assessment and comparison of transferability across multiple source models with different architectures in the context of domain shift, without requiring access to any target labels or source data. Finally, we introduce a Selection, Ensemble, and Adaptation (SEA) framework to address Zoo-MSFDA. 
Through extensive experimentation across various benchmarks, we have demonstrated the effectiveness and robustness of our approach in terms of both transferability estimation and adaptation performance.


%% file: mainbody.bbl
\begin{thebibliography}{10}
\providecommand{\url}[1]{#1}
\csname url@samestyle\endcsname
\providecommand{\newblock}{\relax}
\providecommand{\bibinfo}[2]{#2}
\providecommand{\BIBentrySTDinterwordspacing}{\spaceskip=0pt\relax}
\providecommand{\BIBentryALTinterwordstretchfactor}{4}
\providecommand{\BIBentryALTinterwordspacing}{\spaceskip=\fontdimen2\font plus
\BIBentryALTinterwordstretchfactor\fontdimen3\font minus \fontdimen4\font\relax}
\providecommand{\BIBforeignlanguage}[2]{{%
\expandafter\ifx\csname l@#1\endcsname\relax
\typeout{** WARNING: IEEEtran.bst: No hyphenation pattern has been}%
\typeout{** loaded for the language `#1'. Using the pattern for}%
\typeout{** the default language instead.}%
\else
\language=\csname l@#1\endcsname
\fi
#2}}
\providecommand{\BIBdecl}{\relax}
\BIBdecl

\bibitem{ganin2016domain}
Y.~Ganin, E.~Ustinova, H.~Ajakan, P.~Germain, H.~Larochelle, F.~Laviolette, M.~Marchand, and V.~Lempitsky, ``Domain-adversarial training of neural networks,'' \emph{The journal of machine learning research}, vol.~17, no.~1, pp. 2096--2030, 2016.

\bibitem{shen2021cdtd}
Z.~Shen, M.~Huang, J.~Shi, Z.~Liu, H.~Maheshwari, Y.~Zheng, X.~Xue, M.~Savvides, and T.~S. Huang, ``Cdtd: A large-scale cross-domain benchmark for instance-level image-to-image translation and domain adaptive object detection,'' \emph{International Journal of Computer Vision}, vol. 129, pp. 761--780, 2021.

\bibitem{long2015learning}
M.~Long, Y.~Cao, J.~Wang, and M.~Jordan, ``Learning transferable features with deep adaptation networks,'' in \emph{International conference on machine learning}.\hskip 1em plus 0.5em minus 0.4em\relax PMLR, 2015, pp. 97--105.

\bibitem{ben2010theory}
S.~Ben-David, J.~Blitzer, K.~Crammer, A.~Kulesza, F.~Pereira, and J.~W. Vaughan, ``A theory of learning from different domains,'' \emph{Machine learning}, vol.~79, no.~1, pp. 151--175, 2010.

\bibitem{xu2014guest}
D.~Xu, R.~Chellappa, T.~Darrell, and H.~Daum{\'e}~III, ``Guest editor’s introduction to the special issue on domain adaptation for vision applications,'' \emph{International Journal of Computer Vision}, vol. 109, pp. 1--2, 2014.

\bibitem{xu2022delving}
Y.~Xu, Z.~Jiang, A.~Men, Y.~Liu, and Q.~Chen, ``Delving into the continuous domain adaptation,'' in \emph{Proceedings of the 30th ACM International Conference on Multimedia}, 2022, pp. 6039--6049.

\bibitem{dong2021confident}
J.~Dong, Z.~Fang, A.~Liu, G.~Sun, and T.~Liu, ``Confident anchor-induced multi-source free domain adaptation,'' \emph{Advances in Neural Information Processing Systems}, vol.~34, pp. 2848--2860, 2021.

\bibitem{ahmed2021unsupervised}
S.~M. Ahmed, D.~S. Raychaudhuri, S.~Paul, S.~Oymak, and A.~K. Roy-Chowdhury, ``Unsupervised multi-source domain adaptation without access to source data,'' in \emph{Proceedings of the IEEE/CVF Conference on Computer Vision and Pattern Recognition}, 2021, pp. 10\,103--10\,112.

\bibitem{pei2024evidential}
J.~Pei, A.~Men, Y.~Liu, X.~Zhuang, and Q.~Chen, ``Evidential multi-source-free unsupervised domain adaptation,'' \emph{IEEE Transactions on Pattern Analysis and Machine Intelligence}, 2024.

\bibitem{feng2021kd3a}
H.~Feng, Z.~You, M.~Chen, T.~Zhang, M.~Zhu, F.~Wu, C.~Wu, and W.~Chen, ``Kd3a: Unsupervised multi-source decentralized domain adaptation via knowledge distillation.'' in \emph{ICML}, 2021, pp. 3274--3283.

\bibitem{venkateswara2017deep}
H.~Venkateswara, J.~Eusebio, S.~Chakraborty, and S.~Panchanathan, ``Deep hashing network for unsupervised domain adaptation,'' in \emph{Proceedings of the IEEE conference on computer vision and pattern recognition}, 2017, pp. 5018--5027.

\bibitem{he2016deep}
K.~He, X.~Zhang, S.~Ren, and J.~Sun, ``Deep residual learning for image recognition,'' in \emph{Proceedings of the IEEE conference on computer vision and pattern recognition}, 2016, pp. 770--778.

\bibitem{tan2021efficientnetv2}
M.~Tan and Q.~Le, ``Efficientnetv2: Smaller models and faster training,'' in \emph{International conference on machine learning}.\hskip 1em plus 0.5em minus 0.4em\relax PMLR, 2021, pp. 10\,096--10\,106.

\bibitem{liu2021swin}
Z.~Liu, Y.~Lin, Y.~Cao, H.~Hu, Y.~Wei, Z.~Zhang, S.~Lin, and B.~Guo, ``Swin transformer: Hierarchical vision transformer using shifted windows,'' in \emph{Proceedings of the IEEE/CVF international conference on computer vision}, 2021, pp. 10\,012--10\,022.

\bibitem{mohr2023fast}
F.~Mohr and J.~N. van Rijn, ``Fast and informative model selection using learning curve cross-validation,'' \emph{IEEE Transactions on Pattern Analysis and Machine Intelligence}, 2023.

\bibitem{you2021logme}
K.~You, Y.~Liu, J.~Wang, and M.~Long, ``Logme: Practical assessment of pre-trained models for transfer learning,'' in \emph{International Conference on Machine Learning}.\hskip 1em plus 0.5em minus 0.4em\relax PMLR, 2021, pp. 12\,133--12\,143.

\bibitem{bolya10495258scalable}
D.~Bolya, R.~Mittapalli, and J.~Hoffman, ``Scalable diverse model selection for accessible transfer learning, 2021,'' \emph{ISSN}, vol. 10495258, no.~2, p.~3.

\bibitem{hu2024mixed}
D.~Hu, J.~Liang, J.~H. Liew, C.~Xue, S.~Bai, and X.~Wang, ``Mixed samples as probes for unsupervised model selection in domain adaptation,'' \emph{Advances in Neural Information Processing Systems}, vol.~36, 2024.

\bibitem{bachu2023building}
S.~Bachu, T.~Garg, N.~L. Narasimhan, R.~Konuru, V.~N. Balasubramanian \emph{et~al.}, ``Building a winning team: Selecting source model ensembles using a submodular transferability estimation approach,'' in \emph{Proceedings of the IEEE/CVF International Conference on Computer Vision}, 2023, pp. 11\,609--11\,620.

\bibitem{chen2023explore}
Y.~Chen, T.~Hu, F.~Zhou, Z.~Li, and Z.-M. Ma, ``Explore and exploit the diverse knowledge in model zoo for domain generalization,'' in \emph{International Conference on Machine Learning}.\hskip 1em plus 0.5em minus 0.4em\relax PMLR, 2023, pp. 4623--4640.

\bibitem{bahng2020learning}
H.~Bahng, S.~Chun, S.~Yun, J.~Choo, and S.~J. Oh, ``Learning de-biased representations with biased representations,'' in \emph{International Conference on Machine Learning}.\hskip 1em plus 0.5em minus 0.4em\relax PMLR, 2020, pp. 528--539.

\bibitem{tran2019transferability}
A.~T. Tran, C.~V. Nguyen, and T.~Hassner, ``Transferability and hardness of supervised classification tasks,'' in \emph{Proceedings of the IEEE/CVF International Conference on Computer Vision}, 2019, pp. 1395--1405.

\bibitem{nguyen2020leep}
C.~Nguyen, T.~Hassner, M.~Seeger, and C.~Archambeau, ``Leep: A new measure to evaluate transferability of learned representations,'' in \emph{International Conference on Machine Learning}.\hskip 1em plus 0.5em minus 0.4em\relax PMLR, 2020, pp. 7294--7305.

\bibitem{pei2023uncertainty}
J.~Pei, Z.~Jiang, A.~Men, L.~Chen, Y.~Liu, and Q.~Chen, ``Uncertainty-induced transferability representation for source-free unsupervised domain adaptation,'' \emph{IEEE Transactions on Image Processing}, 2023.

\bibitem{ovadia2019can}
Y.~Ovadia, E.~Fertig, J.~Ren, Z.~Nado, D.~Sculley, S.~Nowozin, J.~Dillon, B.~Lakshminarayanan, and J.~Snoek, ``Can you trust your model's uncertainty? evaluating predictive uncertainty under dataset shift,'' \emph{Advances in neural information processing systems}, vol.~32, 2019.

\bibitem{peng2024energy}
R.~Peng, H.~Zou, H.~Wang, Y.~Zeng, Z.~Huang, and J.~Zhao, ``Energy-based automated model evaluation,'' \emph{arXiv preprint arXiv:2401.12689}, 2024.

\bibitem{yu2022predicting}
Y.~Yu, Z.~Yang, A.~Wei, Y.~Ma, and J.~Steinhardt, ``Predicting out-of-distribution error with the projection norm,'' in \emph{International Conference on Machine Learning}.\hskip 1em plus 0.5em minus 0.4em\relax PMLR, 2022, pp. 25\,721--25\,746.

\bibitem{gretton2006kernel}
A.~Gretton, K.~Borgwardt, M.~Rasch, B.~Sch{\"o}lkopf, and A.~Smola, ``A kernel method for the two-sample-problem,'' \emph{Advances in neural information processing systems}, vol.~19, pp. 513--520, 2006.

\bibitem{ding2022source}
N.~Ding, Y.~Xu, Y.~Tang, C.~Xu, Y.~Wang, and D.~Tao, ``Source-free domain adaptation via distribution estimation,'' in \emph{Proceedings of the IEEE/CVF Conference on Computer Vision and Pattern Recognition}, 2022, pp. 7212--7222.

\bibitem{tian2021source}
Q.~Tian, C.~Ma, F.-Y. Zhang, S.~Peng, and H.~Xue, ``Source-free unsupervised domain adaptation with sample transport learning,'' \emph{Journal of Computer Science and Technology}, vol.~36, no.~3, pp. 606--616, 2021.

\bibitem{liang2020we}
J.~Liang, D.~Hu, and J.~Feng, ``Do we really need to access the source data? source hypothesis transfer for unsupervised domain adaptation,'' in \emph{International Conference on Machine Learning}.\hskip 1em plus 0.5em minus 0.4em\relax PMLR, 2020, pp. 6028--6039.

\bibitem{liang2021source}
J.~Liang, D.~Hu, Y.~Wang, R.~He, and J.~Feng, ``Source data-absent unsupervised domain adaptation through hypothesis transfer and labeling transfer,'' \emph{IEEE Transactions on Pattern Analysis and Machine Intelligence}, 2021.

\bibitem{zhang2022divide}
Z.~Zhang, W.~Chen, H.~Cheng, Z.~Li, S.~Li, L.~Lin, and G.~Li, ``Divide and contrast: Source-free domain adaptation via adaptive contrastive learning,'' \emph{Advances in Neural Information Processing Systems}, vol.~35, pp. 5137--5149, 2022.

\bibitem{yang2021exploiting}
S.~Yang, J.~van~de Weijer, L.~Herranz, S.~Jui \emph{et~al.}, ``Exploiting the intrinsic neighborhood structure for source-free domain adaptation,'' \emph{Advances in neural information processing systems}, vol.~34, pp. 29\,393--29\,405, 2021.

\bibitem{mohr2022lcdb}
F.~Mohr, T.~J. Viering, M.~Loog, and J.~N. van Rijn, ``Lcdb 1.0: An extensive learning curves database for classification tasks,'' in \emph{Joint European Conference on Machine Learning and Knowledge Discovery in Databases}.\hskip 1em plus 0.5em minus 0.4em\relax Springer, 2022, pp. 3--19.

\bibitem{viering2022shape}
T.~Viering and M.~Loog, ``The shape of learning curves: a review,'' \emph{IEEE Transactions on Pattern Analysis and Machine Intelligence}, vol.~45, no.~6, pp. 7799--7819, 2022.

\bibitem{bolya2021scalable}
D.~Bolya, R.~Mittapalli, and J.~Hoffman, ``Scalable diverse model selection for accessible transfer learning,'' \emph{Advances in Neural Information Processing Systems}, vol.~34, pp. 19\,301--19\,312, 2021.

\bibitem{sugiyama2007covariate}
M.~Sugiyama, M.~Krauledat, and K.-R. M{\"u}ller, ``Covariate shift adaptation by importance weighted cross validation.'' \emph{Journal of Machine Learning Research}, vol.~8, no.~5, 2007.

\bibitem{you2019towards}
K.~You, X.~Wang, M.~Long, and M.~Jordan, ``Towards accurate model selection in deep unsupervised domain adaptation,'' in \emph{International Conference on Machine Learning}.\hskip 1em plus 0.5em minus 0.4em\relax PMLR, 2019, pp. 7124--7133.

\bibitem{morerio2017minimal}
P.~Morerio, J.~Cavazza, and V.~Murino, ``Minimal-entropy correlation alignment for unsupervised deep domain adaptation,'' \emph{arXiv preprint arXiv:1711.10288}, 2017.

\bibitem{saito2021tune}
K.~Saito, D.~Kim, P.~Teterwak, S.~Sclaroff, T.~Darrell, and K.~Saenko, ``Tune it the right way: Unsupervised validation of domain adaptation via soft neighborhood density,'' in \emph{Proceedings of the IEEE/CVF international conference on computer vision}, 2021, pp. 9184--9193.

\bibitem{ahmed2021adaptive}
W.~Ahmed, P.~Morerio, and V.~Murino, ``Adaptive pseudo-label refinement by negative ensemble learning for source-free unsupervised domain adaptation,'' \emph{arXiv preprint arXiv:2103.15973}, 2021.

\bibitem{agostinelli2022transferability}
A.~Agostinelli, J.~Uijlings, T.~Mensink, and V.~Ferrari, ``Transferability metrics for selecting source model ensembles,'' in \emph{Proceedings of the IEEE/CVF Conference on Computer Vision and Pattern Recognition}, 2022, pp. 7936--7946.

\bibitem{dong2022zood}
Q.~Dong, A.~Muhammad, F.~Zhou, C.~Xie, T.~Hu, Y.~Yang, S.-H. Bae, and Z.~Li, ``Zood: Exploiting model zoo for out-of-distribution generalization,'' \emph{Advances in Neural Information Processing Systems}, vol.~35, pp. 31\,583--31\,598, 2022.

\bibitem{kundu2020universal}
J.~N. Kundu, N.~Venkat, R.~V. Babu \emph{et~al.}, ``Universal source-free domain adaptation,'' in \emph{Proceedings of the IEEE/CVF Conference on Computer Vision and Pattern Recognition}, 2020, pp. 4544--4553.

\bibitem{vapnik1991principles}
V.~Vapnik, ``Principles of risk minimization for learning theory,'' \emph{Advances in neural information processing systems}, vol.~4, 1991.

\bibitem{shen2023balancing}
M.~Shen, Y.~Bu, and G.~W. Wornell, ``On balancing bias and variance in unsupervised multi-source-free domain adaptation,'' in \emph{International Conference on Machine Learning}.\hskip 1em plus 0.5em minus 0.4em\relax PMLR, 2023, pp. 30\,976--30\,991.

\bibitem{gretton2007kernel}
A.~Gretton, K.~Fukumizu, C.~Teo, L.~Song, B.~Sch{\"o}lkopf, and A.~Smola, ``A kernel statistical test of independence,'' \emph{Advances in neural information processing systems}, vol.~20, 2007.

\bibitem{yang2022attracting}
S.~Yang, Y.~Wang, K.~Wang, S.~Jui \emph{et~al.}, ``Attracting and dispersing: A simple approach for source-free domain adaptation,'' in \emph{Advances in Neural Information Processing Systems}, 2022.

\bibitem{saenko2010adapting}
K.~Saenko, B.~Kulis, M.~Fritz, and T.~Darrell, ``Adapting visual category models to new domains,'' in \emph{European conference on computer vision}.\hskip 1em plus 0.5em minus 0.4em\relax Springer, 2010, pp. 213--226.

\bibitem{peng2019moment}
X.~Peng, Q.~Bai, X.~Xia, Z.~Huang, K.~Saenko, and B.~Wang, ``Moment matching for multi-source domain adaptation,'' in \emph{Proceedings of the IEEE/CVF international conference on computer vision}, 2019, pp. 1406--1415.

\bibitem{huang2022frustratingly}
L.-K. Huang, J.~Huang, Y.~Rong, Q.~Yang, and Y.~Wei, ``Frustratingly easy transferability estimation,'' in \emph{International Conference on Machine Learning}.\hskip 1em plus 0.5em minus 0.4em\relax PMLR, 2022, pp. 9201--9225.

\bibitem{spearman1987proof}
C.~Spearman, ``The proof and measurement of association between two things,'' \emph{The American journal of psychology}, vol. 100, no. 3/4, pp. 441--471, 1987.

\end{thebibliography}
